\documentclass[review, journal]{IEEEtran}

\usepackage{amsfonts}
\usepackage{amsmath}
\usepackage{amsmath,epsfig,amsmath}
\usepackage{amsfonts,amssymb,amsthm}
\usepackage{epsfig}
\usepackage{float}
\usepackage{graphicx}
\usepackage{color,hyperref}
\usepackage{multirow}
\usepackage{tipa}
\usepackage{picins}
\usepackage{color}
\usepackage{booktabs}
\usepackage{threeparttable}

\begin{document}

\title{Gating Syn-to-Real Knowledge for Pedestrian Crossing Prediction in Safe Driving}

\author{Jie Bai, Jianwu Fang, Yisheng Lv, Chen Lv, Jianru Xue, and Zhengguo Li
\thanks{J. Fang and J. Xue are with Xi'an Jiaotong University, Xi'an, China
     {(fangjianwu@mail.xjtu.edu.cn).}}% 
\thanks{J. Bai is with the College of Transportation Engineering, Chang'an University, Xi'an, China.}%   % 
\thanks{Y. Lv is with the Institute of Automation, Chinese Academy of Sciences, Beijing, China}%
        \thanks{C. Lv is with Nanyang Technological University, Singapore}%
 \thanks{Z. Li is with A$^*$STAR, Singapore}

}

\markboth{}%
{Shell \MakeLowercase{\textit{et al.}}: Bare Demo of IEEEtran.cls for Computer Society Journals}

\IEEEtitleabstractindextext{%
\begin{abstract}
Pedestrian Crossing Prediction (PCP) in driving scenes plays a critical role in ensuring the safe operation of intelligent vehicles. Due to the limited observations of pedestrian crossing behaviors in typical situations, recent studies have begun to leverage synthetic data with flexible variation to boost prediction performance, employing domain adaptation frameworks. However, different domain knowledge has distinct cross-domain distribution gaps, which necessitates suitable domain knowledge adaption ways for PCP tasks. In this work, we propose a Gated Syn-to-Real Knowledge transfer approach for PCP (Gated-S2R-PCP), which has two aims: 1) designing the suitable domain adaptation ways for different kinds of crossing-domain knowledge, and 2) transferring suitable knowledge for specific situations with gated knowledge fusion. Specifically, we design a framework that contains three domain adaption methods including style transfer, distribution approximation, and knowledge distillation for various information, such as visual, semantic, depth, location, \emph{etc}. A Learnable Gated Unit (LGU) is employed to fuse suitable cross-domain knowledge to boost pedestrian crossing prediction. We construct a new synthetic benchmark S2R-PCP-3181 with 3181 sequences (489,740 frames) which contains the pedestrian locations, RGB frames, semantic images, and depth images. With the synthetic S2R-PCP-3181, we transfer the knowledge to two real challenging datasets of PIE and JAAD, and superior PCP performance is obtained to the state-of-the-art methods. 
\end{abstract}

\begin{IEEEkeywords}
Pedestrian crossing prediction, safe driving, domain adaptation, gated network, Timesformer
\vspace{4em}
\end{IEEEkeywords}}

\maketitle

\IEEEdisplaynontitleabstractindextext

% \fi
\IEEEpeerreviewmaketitle

\IEEEraisesectionheading{
\section{Introduction}
\label{section1}}
\IEEEPARstart{A}{bout} 50\% road crashes involve vulnerable road users (pedestrians, cyclists, and motorbikes) each year \cite{trafficdeath}. Therefore, safety must be maintained towards automatic or intelligent vehicles, prioritized for the most vulnerable road users \cite{marti2019review}. To enhance safe driving, this work targets the Pedestrian Crossing Prediction (PCP) problem because pedestrian crossing is one of the most typical behaviors that compete for road rights with vehicles \cite{li2021three,DBLP:journals/tiv/RasouliKT18,DBLP:journals/tiv/PredhumeauSD23,vial2020amsense}.

PCP attracts increasing attention with the help of large-scale benchmarks \cite{DBLP:conf/iccvw/RasouliKT17,DBLP:conf/iccv/RasouliKKT19,DBLP:journals/corr/abs-2211-00385}, facilitating various data-driven deep PCP models. Each pedestrian crossing behavior involves various information, such as the pedestrian poses, locations, and pedestrian-to-scene interactions. Additionally, severe weather and light conditions pose challenges further to PCP. Therefore, sample diversity, concerning various states and scene interactions of pedestrians, is vital to boost the generalization of PCP, and borrowing synthetic data is popular for this purpose via domain adaptation models \cite{achaji2022attention,baijie2022,DBLP:journals/ieeejas/MiaoLHWW23,DBLP:journals/itsm/LiuLLCLC22}. This involves the fundamental parallel verification of synthetic and real data for PCP \cite{DBLP:journals/tiv/WangIL22}, whereas the synthetic-real data shift is boring with varying modalities. Naturally, the synthetic (\emph{more information})-to-real (\emph{limited information}) knowledge transfer is straightforward for PCP. The most related ones to our work are \cite{achaji2022attention} and \cite{baijie2022} with a syn-to-real pedestrian location fine-tuning strategy and syn-to-real knowledge distillation framework, respectively. However, the domain gaps are not explicitly investigated. 
 \begin{figure}[!t]
  \centering
 \includegraphics[width=0.97\hsize]{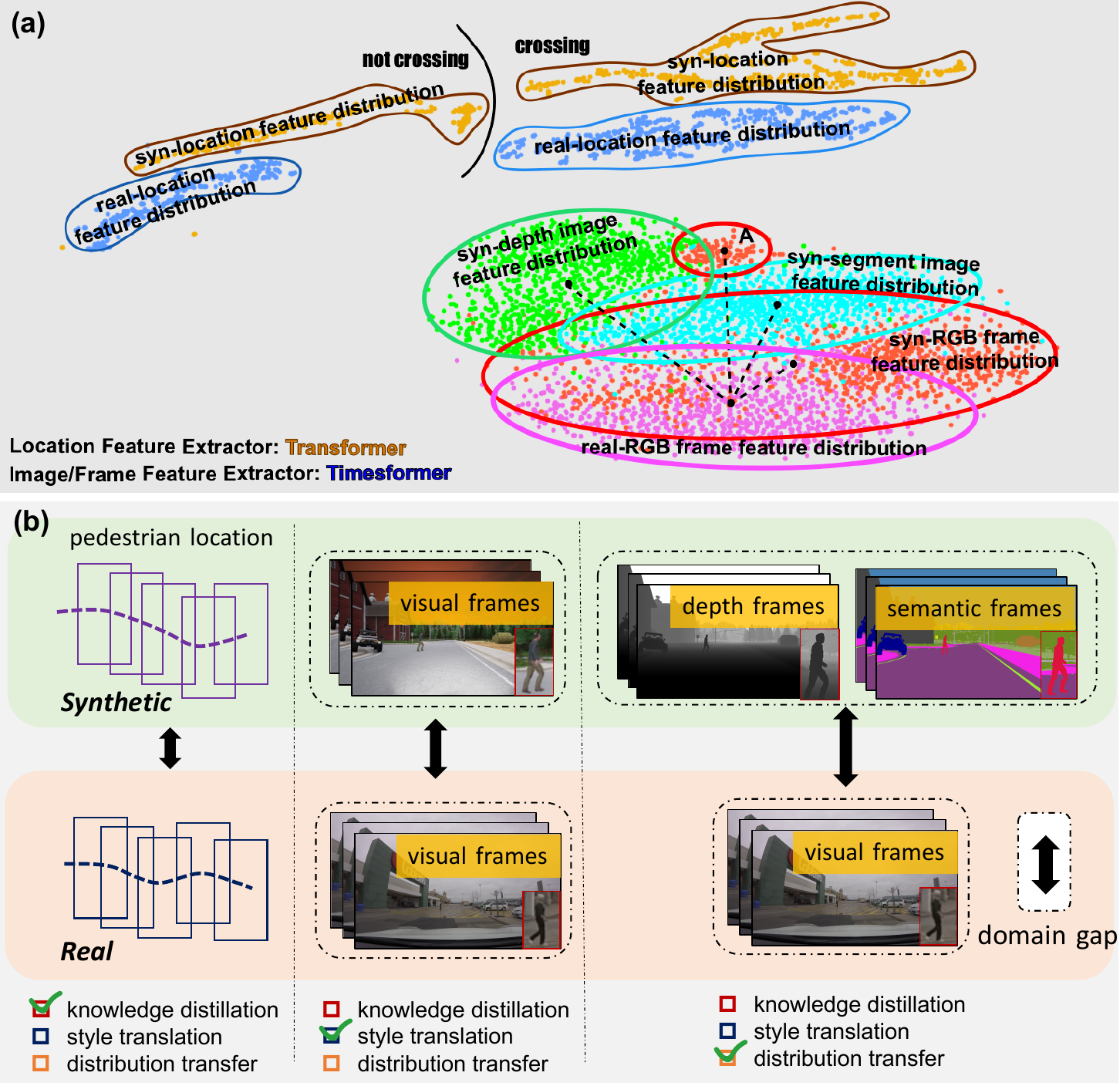}
  \caption{Illustration for the suitable syn-to-real knowledge transfer for different kinds of information in driving scenes. (a) plots the feature distributions of real and synthetic datasets, for the pedestrian locations, RGB frames, and semantic/depth images. (b) illustrates our work for differentiated syn-real knowledge transfer with different domain gaps. The feature distribution in (a) is obtained by t-SNE \cite{van2008visualizing} on the feature vectors of 1000 randomly selected samples (each sample has 16 frames) in the synthetic PCP dataset (Syn-PCP-3181, to be described in Sec. \ref{sec-data}) and a real PCP dataset (PIE \cite{DBLP:conf/iccv/RasouliKKT19}). Notably, because of the different input shapes of pedestrian locations and images, the feature vectors of pedestrian locations and image-like inputs are extracted by Transformer \cite{DBLP:conf/nips/VaswaniSPUJGKP17}  and Timesformer \cite{DBLP:conf/icml/BertasiusWT21}, respectively.}
  \label{fig1}
  \vspace{-0.5em}
\end{figure}

\textbf{Observations}: To illustrate the distinct domain gaps for different information in the PCP task, Fig. \ref{fig1}(a) plots the feature distributions of pedestrian locations, RGB frames, and semantic and depth images. We can observe that the location feature distributions in synthetic and real datasets own a similar manifold and with close intra-distance for crossing or not crossing behaviors. The feature distribution of syn-RGB frames occupies the largest range and correlates with the ones of syn-depth images, syn-segment images, and real RGB frames. Notably, the subset \textbf{A} may be generated by the similar contour or object shapes of the scenes in syn-images (RGB, depth, segment). Based on the observation, the domain gaps for the pairs of $<$syn-loc, real-loc$>$, $<$syn-RGB, real-RGB$>$, and $<$syn-seg/depth, real-RGB$>$ increase gradually. 

\textbf{Solutions}: To address this, we \emph{advocate a differentiated knowledge transfer} for different information within syn-real pairs for adaptive PCP. We propose a Gated Syn-to-Real knowledge transfer model for PCP (Gated-S2R-PCP). As shown in Fig. \ref{fig1}(b), Gated-S2R-PCP advocates three different pathways for transferring the pedestrian locations, visual context, and semantic-depth clues in the synthetic data to the real data that is only with the pedestrian locations and raw RGB frames. Specifically, a Knowledge Distiller, a Style Shifter (StyS), and a Distribution Approximator (DistA) are incorporated for transferring the syn-to-real knowledge in pedestrian location, RGB frames, and semantic-depth images,  respectively. The \textbf{contributions} are threefold. 

\begin{itemize}
\item We propose a gated knowledge transfer framework (Gated-S2R-PCP) that incorporates a Style Shifter, a Distribution Approximator, and a Knowledge Distiller together for better cross-domain learning of the visual context, semantic-depth clue, and pedestrian locations. 
\item We propose a Learnable Gated Unit (LGU) that adaptively learns the weights of different transfer methods to achieve suitable knowledge transfer learning for PCP, to obtain a general knowledge transfer model for different information with distinct synthetic-to-real domain gaps.
\item We construct a large-scale synthetic dataset S2R-PCP-3181\footnote{\url{https://github.com/JWFanggit/Gated-S2R-PCP}}, which contains 3181 sequences with 498,740 frames. S2R-PCP-3181 contains pedestrian locations, RGB frames, depth, and semantic images. Gated-S2R-PCP is verified on JAAD \cite{DBLP:conf/iccvw/RasouliKT17} and PIE \cite{DBLP:conf/iccv/RasouliKKT19} datasets and shows superiority to other SOTA methods. Additionally, we evaluate the Gated-S2R-PCP on all pedestrian crossing sequences (50 ones) in the DADA-2000 dataset \cite{DBLP:journals/tits/FangYQXY22} to verify the PCP performance in near-collision scenes.
\end{itemize}

The remainder of the paper is organized as follows. Sec. \ref{relatework} briefly reviews the related works. Gated-S2R-PCP is described in Sec. \ref{method}, and evaluated by extensive experiments in Sec. \ref{expe}. The conclusion is given in Sec. \ref{con}.

\section{Related Work}
\label{relatework} 
\subsection{Pedestrian Crossing Prediction (PCP)}
Pedestrian crossing prediction in driving scenes can be formulated by classifying the future crossing intention of road participants with a short time observation. Commonly, there is a Time-to-Crossing (TTC) interval ($>0$ seconds) between the end of the observation with the crossing intention. The pedestrian crossing prediction develops from traditional probabilistic models to data-driven deep learning models. 

\textbf{Probabilistic Model-driven Methods:} Because of the high motility of pedestrians, traditional probabilistic model-driven methods model the PCP in the Markov process \cite{DBLP:conf/itsc/HashimotoGHK15} or Bayesian models \cite{DBLP:conf/dagm/SchneiderG13}. In 2013, the Daimler team of Germany and the University of Amsterdam designed a pedestrian path prediction method based on a recursive Bayesian filter \cite{DBLP:conf/itsc/HashimotoGHK15}. Additionally, because of the common moving direction of the pedestrian group, the prediction task of the movement intention of the pedestrian crowd is described as a Markov Decision Process (MDP) problem \cite{DBLP:conf/itsc/HashimotoGHK15,DBLP:journals/tits/ZhuangW13}, which uses social force model and random walk strategy to predict the interactive movement intention of pedestrians. Most traditional probabilistic models adopt the classical probability theory, which is highly interpretable but has limited generalization because of the real scene variability. 

\textbf{Sequential Encoding-Decoding Methods:} With the promotion of deep learning models, most of recent PCP works are formulated as sequential encoding-decoding frameworks. Among them, the Long-Short Term Memory (LSTM) \cite{DBLP:journals/neco/HochreiterS97} and Gated Recurrent Unit (GRU) \cite{dey2017gate} are two universal architectures in sequential information modeling of PCP approaches \cite{DBLP:journals/access/BreharMMVNN21}. Within this field, the pedestrian appearance, posture, trajectory, surrounding scenes, speed, \emph{etc.}, are all investigated by appropriate temporal feature coding and fusion methods \cite{DBLP:conf/bmvc/RasouliKT19,DBLP:conf/iccvw/RasouliKT17,DBLP:conf/iccv/RasouliKKT19,DBLP:conf/ivs/KotserubaRT20,kotseruba2021benchmark}. Among them, the most representative works are the Joint Attention in Autonomous Driving (JAAD) \cite{DBLP:conf/iccvw/RasouliKT17} and Pedestrian Intention Estimation (PIE) \cite{DBLP:conf/iccv/RasouliKKT19}. These two datasets drive the PCP task significantly. Besides LSTM and GRU, 3D convolution is also popular in temporal encoding in the PCP task. For example, Saleh \emph{et al.} \cite{saleh2017intent} adopt a series of 3D DenseNet modules to encode the spatiotemporal information of pedestrians, to judge the real-time crossing intention. However, due to the expensive encoding of 3D convolution, temporal 3D ConvNets (T3D) is introduced by Chaabane \emph{et al.} \cite{chaabane2020looking} for PCP and improves the performance. 

\textbf{Interaction and Attention Driven Models:} Recently, some new deep learning models, such as Graph Convolution Network (GCN) \cite{DBLP:conf/iclr/KipfW17} and Transformer \cite{DBLP:conf/nips/VaswaniSPUJGKP17,DBLP:conf/icra/RasouliK23}, are frequently used in PCP task, with the consideration of the pose, optical flow, semantic context, ego-car velocity, pedestrian locations, and so on. For instance, Neogi \emph{et al.} use the scene context information \cite{DBLP:journals/tits/NeogiHDYD21} in Factored Latent Dynamic Conditional Random Fields (FLDCRF), where PCP is factorized to comprehensively explore the information of pedestrian-vehicle interaction, scene semantics, scene depth, and optical flow, where better results than LSTM modeling is obtained. Cadena \emph{et al.} \cite{20222012127871} extend their work, Pedestrian-Graph \cite{DBLP:conf/itsc/CadenaYQW19}, and propose Pedestrian-Graph+ \cite{20222012127871}. The GCN structure in Pedestrian-Graph+ is changed from \cite{DBLP:conf/itsc/CadenaYQW19} to explore multimodal data information of vehicle speed and image features, respectively. Besides, specific occasions, \emph{e.g.}, the non-signalized intersection \cite{chaabane2020looking} and red-light zone \cite{DBLP:journals/tits/ZhangAWZ22}, are also focused. 

Since the intervention of the Transformer, it has become a popular backbone to attentively encode the information of pedestrians.  PCP with Attention (PCPA) \cite{kotseruba2021benchmark} is a milestone for PCP with attentive modeling for different information. IntFormer \cite{DBLP:journals/corr/abs-2105-08647} generates better PCP performance than PCPA with transformer backbone. Recently, Rasouli \emph{et al.} present the Pedformer \cite{DBLP:conf/icra/RasouliK23} which involves the historical trajectory, ego-car velocity, RGB frames, and frame segments together by a Multiple-Head Attention (MHA) and achieves the highest accuracy (0.93) on the PIE dataset.

\textbf{Discussion}: The aforementioned PCP methods demonstrate significant progress while the variability insufficiency of real datasets makes the existing works involve much more information on pedestrians, ego-cars, and pedestrian-scene interactions to boost the PCP performance, which poses challenges for practical use when some information is not available. In this work, we will explore the domain knowledge transfer method for leveraging various information in the synthetic data to the real data only with pedestrian locations and RGB frames.

\subsection{Cross-Domain Adaptation in Traffic Scenes}
Because of the limited variability of dynamic traffic scenes in one dataset, cross-domain adaptation becomes popular between different datasets (with source datasets and target datasets) \cite{DBLP:conf/wacv/KuznietsovPG22}. In the cross-domain adaption in the driving scene, the core problem is to approximate the distribution between the source domain and target domain. Therefore, distribution alignment \cite{DBLP:journals/tip/JiaoYX21,DBLP:journals/tip/WangLS21}, cross-domain feature distance minimization \cite{DBLP:journals/tits/LiJQ22}, adversarial distribution approximation \cite{lv2018generative,rempe2022generating}, \emph{etc.}, are commonly investigated in this field. For instance, Jiao \emph{et al.} \cite{DBLP:journals/tip/JiaoYX21} propose a Selective Alignment Network (SAN) for cross-domain pedestrian detection. Wang \emph{et al.} \cite{DBLP:journals/tip/WangLS21} model an augmented feature alignment network for cross-domain object detection, where the domain image generation and domain-adversarial training are integrated into a unified framework. Recently, cross-domain object detection has been observed in the autonomous driving community. Li \emph{et al.} \cite{DBLP:journals/tits/LiJQ22} model a stepwise domain adaptation for the popular object detectors and excellent detection performance is obtained. 

In addition to the visual domain adaptation, multi-spectral domain adaptation is another mainstream prototype in the traffic scene \cite{DBLP:journals/cee/ZhangLWCHZL22} to restrict the low visibility conditions issue in RGB frames. For example, Brehar \emph{et al.} \cite{DBLP:journals/access/BreharMMVNN21} explore the role of the far-infrared information of pedestrians in promoting the PCP in low-light conditions, and a CROSSIR dataset is constructed. In this approach, only the far-infrared images are used which does not explore the domain adaptation framework. Marnissi \emph{et al.} \cite{DBLP:journals/prl/MarnissiFSA22} propose a thermal-to-visible domain adaptation for pedestrian detection, where the feature distribution alignments are adopted in the Faster R-CNN. 

With the overview of cross-domain adaption works in traffic scenes, segmentation, and detection are commonly concerned, while the PCP task needs further investigation.
\subsection{Syn-to-Real Knowledge Transfer in Driving Scenes}
Different from the traditional cross-domain adaptation in traffic scenes for real data, syn-to-real knowledge transfer approaches leverage the virtual simulator to generate vast samples based on the traffic layout structure and rules \cite{zhao2022decast}. With the larger diverse weather and light conditions, occasion, and road types, syn-to-real knowledge transfer aims to cover the scene variability and fulfill Parallel Traffic System (PTS) \cite{DBLP:journals/tits/ZhuLCWXW20,DBLP:journals/ieeejas/MiaoLHWW23}. Based on this, researchers in this field begin to construct large-scale datasets with various realistic graph engines (\emph{e.g.}, CARLA \cite{DBLP:conf/corl/DosovitskiyRCLK17}, UE4, \emph{etc.}) for simulating the virtual traffic scene, and adopt the synthetic data in detection, segmentation, and tracking \cite{DBLP:conf/cvpr/SunSPWGSTY22,DBLP:journals/tiv/WangSTWGWYS22}. For instance, Wang \emph{et al.} \cite{DBLP:journals/tiv/WangSTWGWYS22} contribute a parallel teacher for syn-to-real domain adaptation in traffic object detection, which consists of the data level and knowledge level domain distribution shift and takes the knowledge in the real world and virtual world together to achieve the knowledge distillation. As for the PCP task, the syn-to-real knowledge transfer emerged recently \cite{achaji2022attention,baijie2022}, such as the syn-to-real fine-tuning for bounding box information learning \cite{achaji2022attention} and syn-to-real knowledge distillation \cite{baijie2022} involved by ourselves. 

\textbf{Discussion}: Building upon the state-of-the-art, similar to previous PCP works, we advocate involving multiple information of pedestrians and road scenes. Differently, we encode various information in the synthetic data and prefer a PCP in the real datasets only with pedestrian locations and RGB frames. Additionally, different from previous domain knowledge transfer frameworks with a single type of fine-tuning, distillation, or distribution transfer ways, we involve multiple domain knowledge through distinct approaches, such as distillation, style-shifting, and distribution-approximation, to adapt to the differentiated knowledge transfer within different syn-real information pairs.
  
\section{Approach}
\label{method}

This section describes the differentiated knowledge transfer approach of different information in the PCP problem. 
\begin{figure*}[!t]
\centering
\includegraphics[width=\hsize]{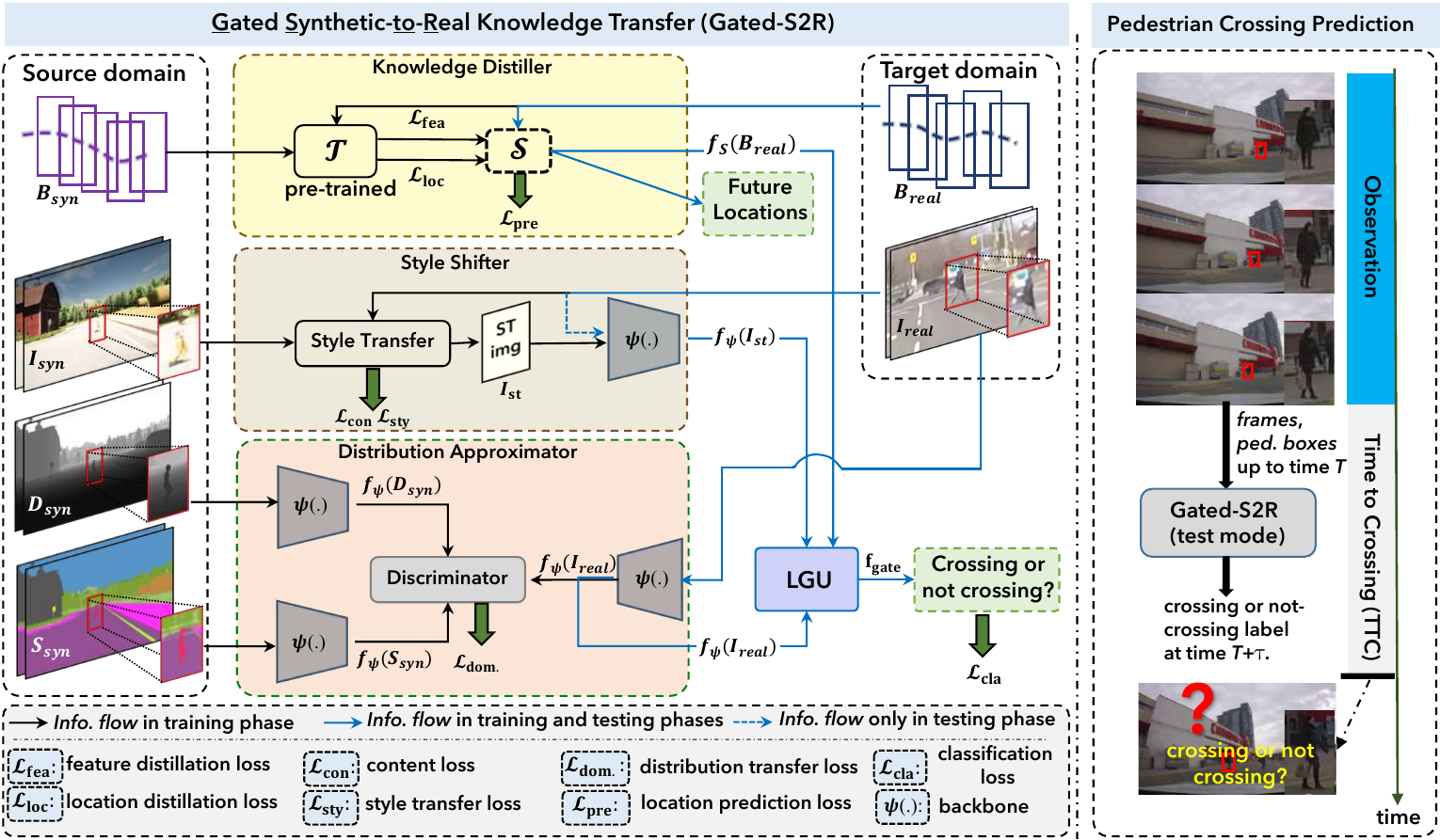}
\caption{The framework of \textbf{Gated-S2R-PCP}. \textbf{Synthetic domain:} pedestrian bounding boxes $\emph{\textbf{B}}_{syn}$, RGB frames $\emph{\textbf{I}}_{syn}$, depth images $\emph{\textbf{D}}_{syn}$ and semantic images $\emph{\textbf{S}}_{syn}$. \textbf{Real domain:} pedestrian bounding boxes $\emph{\textbf{B}}_{real}$ and RGB frames $\emph{\textbf{I}}_{real}$. During the training phase, the Knowledge Distiller, Style Shifer, and Distribution Approximator are trained to fulfill the knowledge transfer of $\emph{\textbf{B}}_{syn}\rightarrow \emph{B}_{real}$ (Eq. \ref{eq:KD}), $\emph{\textbf{I}}_{syn}\rightarrow \emph{\textbf{I}}_{real}$ (Eq. \ref{eq:SS}), and $\{\emph{\textbf{D}}_{syn},\emph{\textbf{S}}_{syn}\}\longleftrightarrow \emph{\textbf{I}}_{real}$ (Eq. \ref{eq:DA}), respectively. With these domain adaptation approaches, we obtain the feature embedding of $f_{\mathcal{S}}(\emph{\textbf{B}}_{real})$ of $\emph{\textbf{B}}_{real}$, $f_{\psi}(\emph{\textbf{I}}_{st})$ for style-transferred image set $\emph{\textbf{I}}_{st}$ and the $f_{\psi}(\emph{\textbf{I}}_{real})$ for $\emph{\textbf{I}}_{real}$ over $T$ frames. Finally, a Learnable Gated Unit (LGU) adaptively fuses $f_{\mathcal{S}}(\emph{\textbf{B}}_{real})$, $f_{\psi}(\emph{\textbf{I}}_{st})$ and $f_{\psi}(\emph{\textbf{I}}_{real})$ to form the multi-source feature $f_{gate}$ for pedestrian crossing prediction. During the testing phase, only $\emph{\textbf{I}}_{real}$ and $\emph{\textbf{B}}_{real}$ are the input up to time $T$ (1:$T$). The pedestrian crossing predictor determines the crossing or not crossing label of the pedestrians at time $T+\tau$, where $\tau$ means the Time-to-Crossing (TTC).}
\label{fig2}
\end{figure*}

\textbf{Problem Formulation.} Assume we have the short observation with $T$ video frames, the PCP aims to predict the crossing or not-crossing label at time $T+\tau$, where $\tau>0$ means the Time-to-Crossing (TTC) interval from time $T$ to $T+\tau$. We assume that there are four modalities (RGB frames, ped. boxes, depth images, and semantic images) and two modalities (only with RGB frames and ped. boxes) in synthetic $\mathcal{V}_{syn}=\{\emph{\textbf{I}}_{syn},\emph{\textbf{B}}_{syn},\emph{\textbf{D}}_{syn}, \emph{\textbf{S}}_{syn}\}$ and real $\mathcal{V}_{real}=\{\emph{\textbf{I}}_{real}, \emph{\textbf{B}}_{real}\}$ datasets, respectively. As defined by Eq. \ref{eq:1}, PCP can be formulated as the following inference problem
\begin{equation}
{\hat{\bf{p}}}_{T+\tau}=G(\phi(\mathcal{V}_{real}|_{1:T},{\bf{W}})),
\label{eq:1}
\end{equation}
where $G$ is the pedestrian crossing determiner with the input of \emph{ped. boxes} and \emph{RGB frames} of the $T$ frames in real datasets, and $\phi(.,.)$ is the feature encoder for $\mathcal{V}_{real}$. $\hat{\bf{p}}_{T+\tau}=\{c,nc\}$ denotes the predicted label of crossing or not-crossing. ${\bf{W}}$ is the model parameters in $\phi(.,.)$, which is obtained by training the knowledge transfer models from synthetic dataset to real dataset. Since the modalities of synthetic and real datasets are different, the knowledge transfer models are formulated by the degree of domain gaps. 

The framework of the Gated-S2R-PCP model is presented in Fig. \ref{fig2}. For the ``\emph{ped. boxes}", we formulate a \textbf{Knowledge Distiller (KnowD)} with a teacher model and a student model for $\emph{\textbf{B}}_{syn}$ and $\emph{\textbf{B}}_{real}$, respectively. As for the RGB frames, because of the same modality, the main difference is the image styles, and we model a \textbf{Style Shifter (StyS)} to transfer the diverse weather and light conditions in $\emph{\textbf{I}}_{syn}$ to $\emph{\textbf{I}}_{real}$. For the depth and semantic images, they have the largest distribution gap to RGB frames, we design a \textbf{Distribution Approximator (DistA)} to project the feature embedding of different modalities into a \emph{shared} feature space for the PCP task. Consequently, we own a KnowD, a StyS, and a DistA to fulfill the differentiated knowledge transfer. 
To achieve an adaptive domain knowledge selection, we fuse different knowledge transfer functions by a Learnable Gated Unit (LGU) to fulfill an ensemble knowledge transfer learning in an end-to-end mode, as defined by Eq. \ref{eq:2}.
\begin{equation}
\begin{array}{ll} 
{f_{gate}}=\mathop{LGU}\limits_{{\bf{W}}_{LGU}}(\mathop{KnowD}\limits_{{\bf{W}}_K}(\emph{\textbf{B}}_{syn}\rightarrow \emph{\textbf{B}}_{real}),\\
\verb'     '\mathop{StyS}\limits_{{\bf{W}}_S}(\emph{\textbf{I}}_{syn}\rightarrow \emph{\textbf{I}}_{real}),\\
\verb'     '\mathop{DistA}\limits_{{\bf{W}}_D}(\{\emph{\textbf{D}}_{syn}, \emph{\textbf{S}}_{syn}\}\longleftrightarrow \emph{\textbf{I}}_{real})),
\end{array} 
\label{eq:2}
\end{equation}
where ${\bf{W}}_K,{\bf{W}}_S$, ${\bf{W}}_D$, and ${\bf{W}}_{LGU}$ are the weights in $\mathop{KnowD}(\rightarrow$), $\mathop{StyS}(\rightarrow$), $\mathop{DistA}(\longleftrightarrow$), and LGU, respectively. ${f_{gate}}$ is the feature vector for PCP task.

\subsection{Knowledge Distiller (KnowD)}
\label{sec:kd}
KnowD contains a teacher model and a student model for pedestrian location learning of $\emph{\textbf{B}}_{syn}$ and $\emph{\textbf{B}}_{real}$, respectively. 
\begin{figure}[!t]
\centering
\includegraphics[width=\hsize]{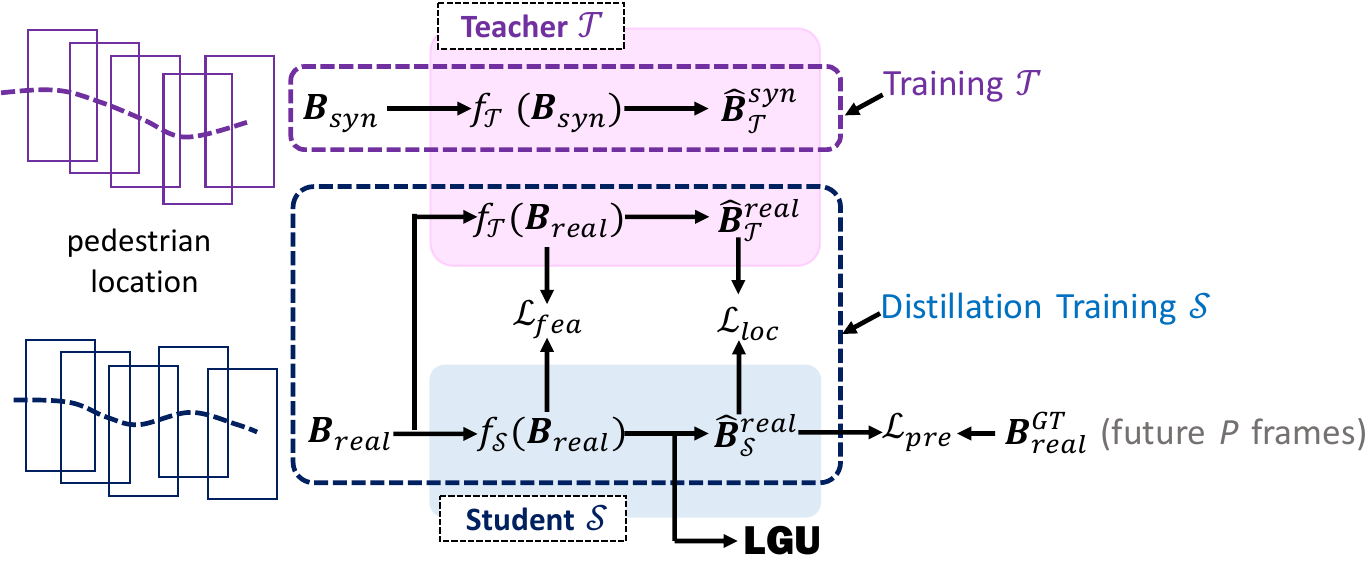}
\caption{The structure of Knowledge Distiller.}
\label{fig3}
\end{figure}
Suppose the pedestrian location $\emph{\textbf{B}}$ be $\{{\bf{b}}_t\}_{t=1}^T$, where ${\bf{b}}_t=(x_t, y_t, w_t, h_t)$ represents the center point ($x_t,y_t$), width, and height of the bounding box, respectively.

In KnowD, the teacher model $\mathcal{T}$ uses synthetic location data to learn the pedestrian behavior pattern, and the pre-trained $\mathcal{T}$ is then adopted to guide the student model learning $\mathcal{S}$ in real location data. As formulated by Eq. \ref{eq:KD},  KnowD is defined as:
\begin{equation}
f_{\mathcal{S}}(\emph{\textbf{B}}_{\emph{real}})=\mathop{KnowD}(\mathcal{T}(\emph{\textbf{B}}_{syn})\rightarrow \mathcal{S}(\emph{\textbf{B}}_{real}),{\bf{W}}_K),
\label{eq:KD}
\end{equation}
where $f_{\mathcal{S}}(\emph{\textbf{B}}_{\emph{real}})$ represents the feature representation after the distillation learning process, which is utilized for subsequent pedestrian crossing prediction task. Fig. \ref{fig3} presents the structure of KnowD. In addition to the feature representation learning, we incorporate a prediction path of the Future Object Location  (\textbf{FOL}) to improve the PCP performance.

Consequently, given $T$ successive frames, the output is the feature representation of pedestrian locations of $T$ successive frames for the PCP and pedestrian locations in future $P$ frames of the FOL, \emph{i.e.}, predicting the locations at the time window $[T+1:T+P]$, where $P$ is evaluated in the experiments. 

\textbf{Training $\mathcal{T}$:} We adopt the Transformer network \cite{DBLP:conf/nips/VaswaniSPUJGKP17} as the backbone model of $\mathcal{T}$. Firstly, the pedestrian locations $\emph{\textbf{B}}_{\emph{syn}}=\{{\bf{b}}_t^{\emph{syn}}\}_{t=1}^{T}$ in the synthetic dataset, is the input of the teacher model $\mathcal{T}$, and $\hat{\emph{\textbf{B}}}_{\mathcal{T}}^{syn}=\{{\hat{\bf{b}}}_t\}_{t=T+1}^{T+P}$ is the pedestrian locations to be predicted. $\emph{\textbf{B}}_{\emph{syn}}$ is first encoded by $\mathcal{T}$ to a feature embedding by Position Embedding Layer in Transformer. It is then encoded by Transformer (with seven interleaved Multi-Head Attention (MHA) and Linear Normalization (LN)) as a 512-dimension feature vector ${\bf{\emph{f}}}_{\mathcal{T}}(\emph{\textbf{B}}_{\emph{syn}})$. The feature vector can be used to classify pedestrian crossing labels at time $T+\tau$ and predict $\hat{\emph{\textbf{B}}}_{\mathcal{T}}^{syn}$ in the synthetic dataset. Once $\mathcal{T}$ is trained, the parameters in $\mathcal{T}$ are fixed in the following distillation process.

\textbf{Distillation Training $\mathcal{S}$:} For the student model $\mathcal{S}$, we prefer a lightweight but efficient architecture. In this work, we take the ResNet+LSTM to encode the pedestrian locations on real dataset, where $\emph{\textbf{B}}_{\emph{real}}=\{{\bf{b}}_t^{\emph{real}}\}_{t=1}^{T}$ and $\hat{\emph{\textbf{B}}}_{\mathcal{S}}^{real}=\{{\hat{\bf{b}}}_t^{\emph{real}}\}_{t=T+1}^{T+P}$ are the input and output of $\mathcal{S}$ with the same format as $\mathcal{T}$. To be lightweight, ResNet18 \cite{DBLP:conf/cvpr/HeZRS16} is chosen as the backbone model, and LSTM temporally associates the feature embedding of ResNet18 for the FOL of pedestrians. The feature vector $f_{\mathcal{S}}(\emph{\textbf{B}}_\emph{{real}})$ with dimension 512 is obtained. To achieve effective distillation, we model three types of losses.

1) \emph{Feature Distribution Consistency} ($\mathcal{L}_{fea}$) advocates the shared label for crossing or not-crossing determined by $\mathcal{T}$ and $\mathcal{S}$, and computed by the Kullback–Leibler Divergence (KLD) $\text{KLD}(f_{\mathcal{T}}(\emph{\textbf{B}}_{real}),f_{\mathcal{S}}(\emph{\textbf{B}}_{real}))$.
 
 2) \emph{FOL Consistency} ($\mathcal{L}_{loc}$) expects the equivalent future object locations predicted by $\mathcal{T}$ and $\mathcal{S}$, specified by Mean Square Error, \emph{i.e.},  $\text{MSE}(\hat{\emph{\textbf{B}}}_{\mathcal{S}}^{real},\hat{\emph{\textbf{B}}}_{\mathcal{T}}^{real})$.
 
 3) \emph{FOL Accuracy} ($\mathcal{L}_{pre}$) prefers a minimum location distance between $\hat{\emph{\textbf{B}}}_{\mathcal{S}}^{real}$ and the ground truth $\emph{\textbf{B}}^{GT}_{real}$, and calculated by $\text{MSE}(\hat{{\emph{\textbf{B}}}}_{\mathcal{S}}^{real},\emph{\textbf{B}}^{GT}_{real})$, where $\emph{\textbf{B}}^{GT}_{real}$ represents the ground truth for the pedestrian locations in the time window [$T+1,T+P$]. 
 
 In total, the loss functions in KnowD are summarized as
\begin{equation}
\mathcal{L}_{kd} =\mathcal{L}_{fea}+\mathcal{L}_{loc}+\mathcal{L}_{pre}.
\end{equation}
\subsection{Style Shifter (StyS)}
\label{sec:ss}
\begin{figure}[!t]
\centering
\includegraphics[width=0.7\hsize]{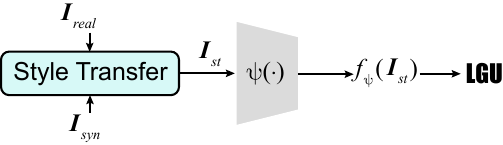}
\caption{The structure of Style Shifter. The real RGB frames $\emph{\textbf{I}}_{real}$ are the content images and the synthetic RGB frames $\emph{\textbf{I}}_{syn}$ are the style images.}
\label{fig4}
\end{figure}

StyS is used to transfer the frame appearance of synthetic RGB frames $\emph{\textbf{I}}_{syn}$ under different conditions to the real RGB frames $\emph{\textbf{I}}_{real}$. This module can be viewed as a scene appearance augmentation for the real dataset. We use the synthetic RGB frames as the style image set and the real RGB frame as the content image set. Then, we formulate StyS using Eq. \ref{eq:SS}, \emph{i.e.}, 
\begin{equation}
f_{\psi}(\emph{\textbf{I}}_{st})=\mathop{StyS}(\emph{\textbf{I}}_{syn}\rightarrow \emph{\textbf{I}}_{real},{\bf{W}}_S),
\label{eq:SS}
\end{equation}
where $f_{\psi}(\emph{\textbf{I}}_{st})$ is the feature embedding of the style-transferred image $\emph{\textbf{I}}_{st}$. The structure of StyS is shown in Fig. \ref{fig4}.

 In this module, both synthetic and real RGB frames stand at the local regions of pedestrians. The global image contains much irrelevant background clutter, which may degrade the following PCP task and cause a slow convergence. We crop the rectangle region with $\beta$ (1.5 in this work) times the size around the pedestrian bounding box, and the cropped regions are scaled with the same size. Consequently, the input region size for StyS is $(T, 3, w, h)$, where [$w, h$] is the region size.

\textbf{Feature-Level Style Transfer}: We advocate a feature-level style transfer to boost the style diversity in StyS. Here, we introduce the efficient AdaIN method \cite{DBLP:conf/iccv/HuangB17} to generate the style-transferred image set $\emph{\textbf{I}}_{st}$ of $T$ frames. Then, $\emph{\textbf{I}}_{st}$ is encoded by a spatial-temporal backbone model $\psi(.)$, and the encoded feature is then fed into the PCP classifier. In this work, different backbones of $\psi(.)$ are experimentally evaluated.
 
To measure the quality of generated $\emph{\textbf{I}}_{st}$, it is necessary to calculate the content loss $\mathcal{L}_{con}$ and style loss $\mathcal{L}_{sty}$. Similarly, the MSE loss between the feature embedding $f_{\psi}(\emph{\textbf{I}}_{real})$ and $f_{\psi}(\emph{\textbf{I}}_{st})$ is adopted to compute the $\mathcal{L}_{con}$. $\mathcal{L}_{sty}$ calculates the MSE of the vector mean and variance of $f_{\psi}(\emph{\textbf{I}}_{syn})$ and $f_{\psi}(\emph{\textbf{I}}_{st})$. Therefore, the total loss in StyS is
\begin{equation}
\mathcal{L}_{st} =\mathcal{L}_{con}+\alpha{\mathcal{L}_{sty}},
\end{equation}
where $\alpha$ is a hyper-parameter and set as $10$ in this work.

\subsection{Distribution Approximator (DistA)}
\label{sec:da}
As the largest domain gap between the synthetic depth images ($\emph{\textbf{D}}_{syn}$), synthetic semantic images ($\emph{\textbf{S}}_{syn}$), and real RGB frames ($\emph{\textbf{I}}_{real}$), we model the DistA by Eq. \ref{eq:DA} to find the shared feature distribution, \emph{i.e.,}
\begin{equation}
f_{\psi}(\emph{\textbf{I}}_{real})=\mathop{DisA}(\{\emph{\textbf{D}}_{syn},\emph{\textbf{S}}_{syn}\}\longleftrightarrow \emph{\textbf{I}}_{real},{\bf{W}}_D).
\label{eq:DA}
\end{equation}
Different from KnowD and StyS, DisA enforces a Bi-directional Approximation.

\begin{figure}[!t]
\centering
\includegraphics[width=\hsize]{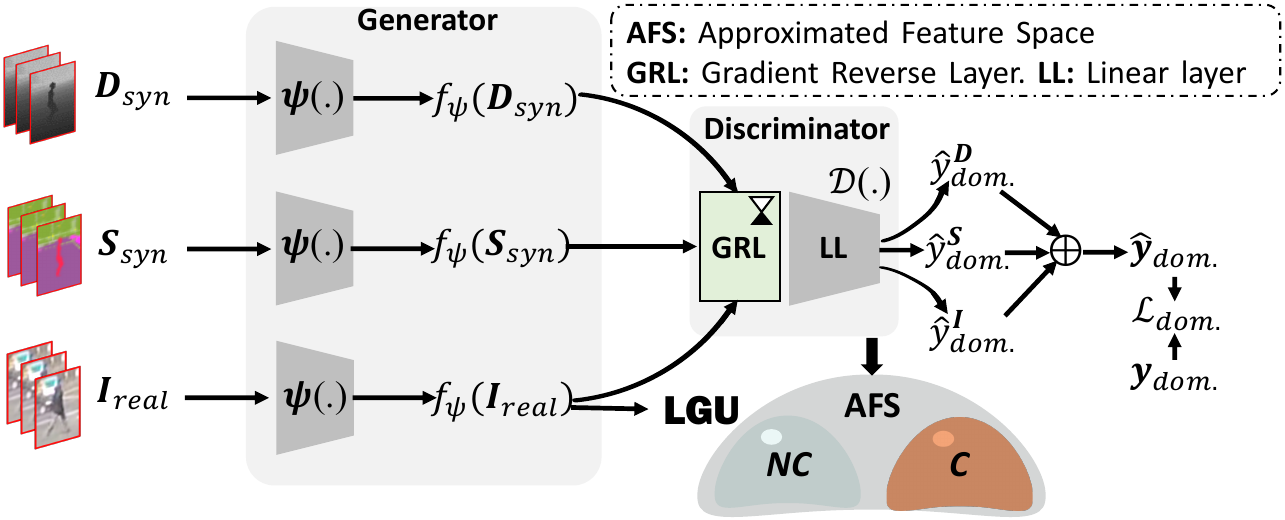}
\caption{The structure of Distribution Approximator. The generator encodes the input to the high-dimensional feature embedding, and the discriminator acts as the domain classifier, which aims to confuse the domain label and approximate the feature embedding of different domains to a shared space.}
\label{fig5}
\end{figure}

DisA is inspired by the adversarial insights of Generative Adversarial Networks (GAN) \cite{DBLP:conf/nips/GoodfellowPMXWOCB14} with the generator-discriminator structure. Differently, DisA focuses on feature-level adversarial training. The generator encodes the $\emph{\textbf{D}}_{syn}$, $\emph{\textbf{S}}_{syn}$, and $\emph{\textbf{I}}_{real}$ by the same backbone model $\psi(.)$ of Style Shifter, which results in the feature embedding of $f_{\psi}(\emph{\textbf{D}}_{syn})$, $f_{\psi}(\emph{\textbf{S}}_{syn})$, and $f_{\psi}(\emph{\textbf{I}}_{real})$, respectively. The discriminator determines whether the feature embedding comes from the synthetic or real domain. The discriminator expects to be confused by the generator for different domains as much as possible while avoiding the GAN-type training. A Gradient Reverse Layer (GRLayer) in \cite{DBLP:conf/icml/GaninL15} is used to reverse the feature space gradient and promote the optimization of the discriminator during training, minimizing the cross-domain feature gap, as shown in Fig. \ref{fig5}. 

In DisA, the data form is the same as the input of StyS, \emph{i.e.}, rectangle regions around the pedestrian boxes. All input data is processed with the same dimensions $(T, c, w, h)$, where channel $c$ equals 1 for $\emph{\textbf{D}}_{syn}$ and $\emph{\textbf{S}}_{syn}$, while $c$ is 3 for $\emph{\textbf{I}}_{real}$.
The spatial regions over $T$ frames are fed into the generator $\psi(.)$ to obtain the feature embedding $f_\psi(\mathcal{Z})$, where $\mathcal{Z}$ denotes the $\emph{\textbf{D}}_{syn}$, $\emph{\textbf{S}}_{syn}$, or $\emph{\textbf{I}}_{real}$.

The discriminator $\mathcal{D}(.)$ consists of the Gradient Reverse Layer and a linear layer. The feature embedding is classified after gradient inversion operation in synthetic and real data domain determination. Therefore, the discriminator can be treated as a domain classifier  (\emph{i.e.}, determining $\hat{y}_{dom.}^{\textbf{\emph{D}}}$, $\hat{y}_{dom.}^{\textbf{\emph{S}}}$, and $\hat{y}_{dom.}^{\textbf{\emph{I}}}$), and the distribution loss is defined by Eq. \ref{eq:8}, \emph{i.e.}, the Binary Cross-Entropy (BCE).
\begin{equation}
\mathcal{L}_{dom.} = \sum_{\mathcal{Z}\in\{\emph{\textbf{D}}_{syn}, \emph{\textbf{S}}_{syn}, \emph{\textbf{I}}_{real}\}} \text{BCE}(\mathcal{D}(f_{\psi}(\mathcal{Z})),{\bf{y}}_{dom.}),
\label{eq:8}
\end{equation}
where ${\bf{y}}_{dom.}=\{0, 1\}$ is the synthetic and real domain labels.

\subsection{Learnable Gated Unit}
\begin{figure}[!t]
\centering
\includegraphics[width=\hsize]{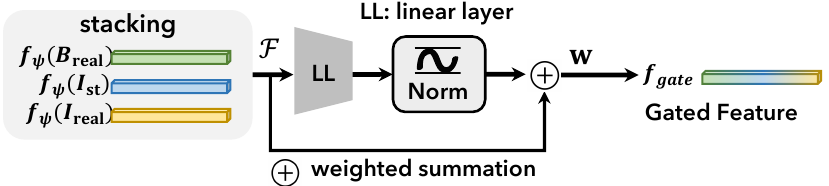}
\caption{The structure of LGU.}
\label{fig6}
\end{figure}

Since the knowledge importance may be different for different scenarios, it is desired to properly set weights for different knowledge transfer functions accordingly. To fulfill an end-to-end ensemble learning, we model a Learnable Gated Unit (LGU) (shown in Fig. \ref{fig6}) to adaptively fuse $f_{\mathcal{S}}(\emph{\textbf{B}}_{real})$ (Eq. \ref{eq:KD}), $f_{\psi}(\emph{\textbf{I}}_{st})$ (Eq. \ref{eq:SS}), and $f_{\psi}(\emph{\textbf{I}}_{real})$ (Eq. \ref{eq:DA}) over $T$ frames. Hence, Eq. \ref{eq:2} is calculated by Eq. \ref{eq:9}.
\begin{equation}
\begin{array}{ll} 
{\mathcal{F}}=[f_{\mathcal{S}}(\emph{\textbf{B}}_{real});f_{\psi}(\emph{\textbf{I}}_{st});f_{\psi}(\emph{\textbf{I}}_{real})],\\
{\bf{w}} = \text{Norm}(\text{LL}({\mathcal{F}},{\bf{W}}_{LGU})),\\
f_{gate} = {\bf{w}} \oplus {\mathcal{F}},
\end{array}
\label{eq:9}
\end{equation}
where [;] denotes a stacking operation for three kinds of features, ${\bf{W}}_{LGU}$ denotes the parameters in LGU, and ${\bf{w}}\in \mathop{R}^{1\times3}$ is the gated weight for feature fusion. LL(.,.) consists of 3 linear layers and a normalization operation. The gate operation is fulfilled by a weighted summation $\oplus$ of ${\mathcal{F}}$ by ${\bf{w}}$.
\subsection{Pedestrian Crossing Prediction}
Thus, the PCP problem in Eq. \ref{eq:1} is re-formulated as
\begin{equation}
{\hat{\bf{p}}}_{T+\tau}=G(f_{gate}),\\
\end{equation}
where $G(.)$ contains 3 linear layers $l(.)$ and a Gumbel-softmax function \cite{DBLP:conf/iclr/JangGP17} which is defined as
\begin{equation}
\sigma_t = \frac{e^{\frac{1}{\eta}{(l(f^i_{gate}) -\log(-\log(\epsilon(i))))}}}{\sum_{j}e^{\frac{1}{\eta}{(l(f^j_{gate}) -\log(-\log(\epsilon(j))))}}},
\end{equation}
where $\sigma_t$ is the output of Gumbel-softmax operation on $l(f_{gate})$, and the logit score of the $i^{th}$ feature item of $l(f_{gate})$ is denoted as $l(f^i_{gate}) -\log(-\log(\epsilon(i)))$, and $\epsilon(i)$ is a random variable that obeys the uniform distribution $U(0,1)$. $l(f_{gate})$ is the output of linear layers $ l(.)$. $\eta$ is the temperature parameter and is set as 1. $G(.)$ is optimized with the BCE loss as
\begin{equation}
\begin{array}{ll}
\mathcal{L}_{cla} =\sum_k(-{\hat{\bf{p}}}_{T+\tau}(k)\log({\bf{p}}_{T+\tau}(k))+\\
\verb'     '(1-{\hat{\bf{p}}}_{T+\tau}(k))\log(1-{\bf{p}}_{T+\tau}(k))),
\end{array}
\end{equation}
 where ${\hat{\bf{p}}}_{T+\tau}(k)$ and ${\bf{p}}_{T+\tau}(k)$ are the predicted and ground-truth intention label $k$ of crossing or not-crossing.
 
The overall loss function of the Gated-S2R-PCP is
\begin{equation}
\mathop{min}\limits_{\{{\bf{W}}_{K},{\bf{W}}_{S},{\bf{W}}_{D},{\bf{W}}_{LGU}\}}: \mathcal{L} = \mathcal{L}_{st}+\mathcal{L}_{dom.}+\mathcal{L}_{kd}+\mathcal{L}_{cla}.\\
\end{equation}
The trained model weights are then used to generate the gated feature $f_{gate}$ for PCP in the testing phase.

\section{Experiments and Discussions}
\label{expe}
\subsection{Datasets}
\label{sec-data}

\begin{figure}[!t]
\centering
\includegraphics[width=\hsize]{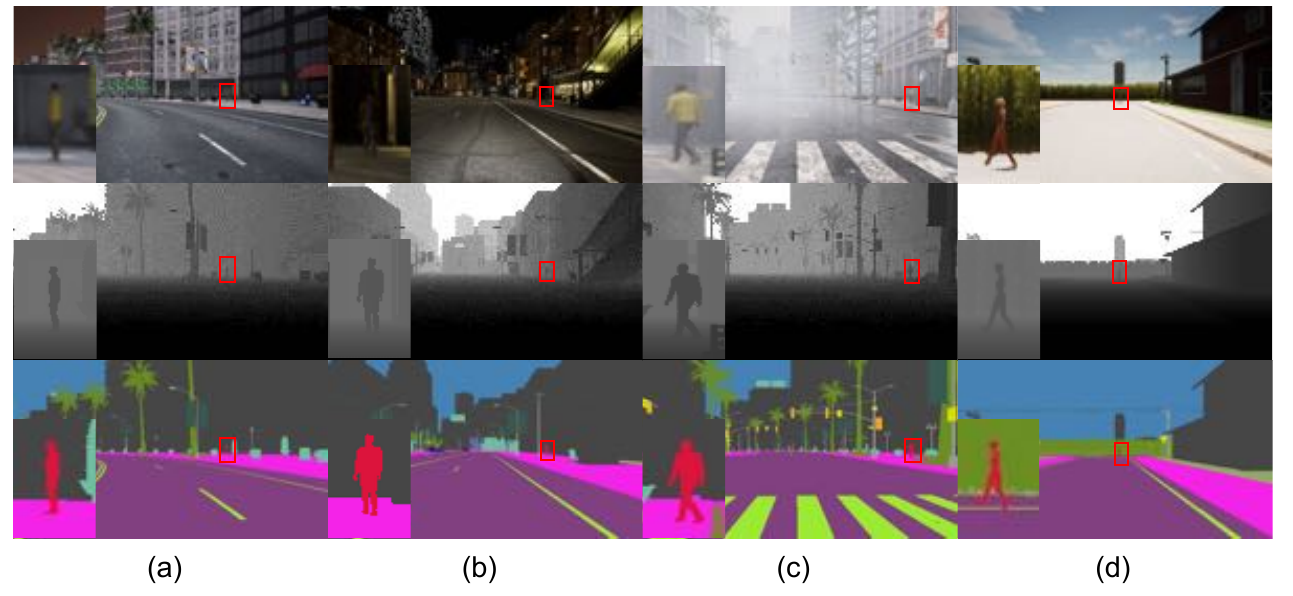}
\caption{Several scenarios in Syn-PCP-3181 dataset. (a), (b), (c) and (d) are the RGB-Depth-Semantic images collected in the urban (daytime), urban (nighttime), rainy, and rural scenes, respectively.}
\label{fig7}
\end{figure}

\begin{figure}[!t]
\centering
\includegraphics[width=0.9\hsize]{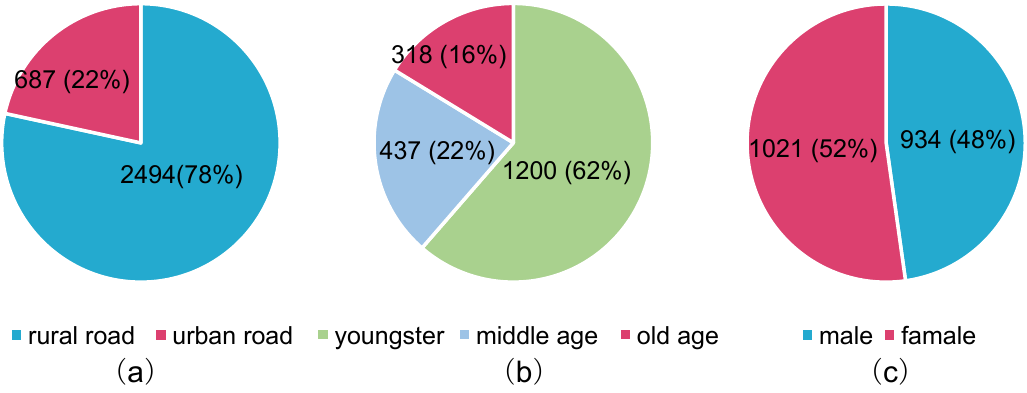}
\caption{The statistics of Syn-PCP-3181 for occasions, gender, and ages.}
\label{fig8}
\end{figure}

\textbf{Synthetic PCP dataset:} In this work, we construct a new synthetic PCP dataset, named Syn-PCP-3181. Syn-PCP-3181 contains 3,181 video sequences with more modalities including RGB frames, depth images, semantic images, and pedestrian locations. Fig. \ref{fig7} displays several examples in Syn-PCP-3181.  
Syn-PCP-3181 is generated by CARLA \cite{dosovitskiy2017carla} under various weather and light conditions, where 1,226 sequences contain crossing behavior and 1955 non-crossing ones. When the dataset is generated, the camera selects the vehicle camera to record the behavior of the pedestrian in front of the vehicle, and the resolution is set to $1280\times720$. To obtain the pedestrian crossing behavior, instead of random setting, we manually set different initial positions and various destination positions in varying urban scenes. To ensure the diversity of the dataset, we also assign different ages and genders to the pedestrians, such as youngest, old age, and middle age. In addition, Syn-PCP-3181 includes a variety of weather (sunny, cloudy, and rainy), light (sunset, daytime, and nighttime), and road occasions (urban, rural). Each kind of weather and light condition takes 1/3 in balance. The occasions, gender, and age statistics are shown in Fig. \ref{fig8}.

To avoid behavior confusion, each video sequence only contains one pedestrian. Therefore, the number of pedestrians is 3,181. Each crossing sequence contains 240 frames, while one not-crossing sequence contains 100 frames, so the total number of frames in the data set can reach 489,740. The most similar dataset to Syn-PCP-3181 is the Virtual-PedCross-4667 \cite{baijie2022} collected by ourselves which only contains the RGB frames and pedestrian locations.

\begin{figure}[!t]
\centering
\includegraphics[width=\hsize]{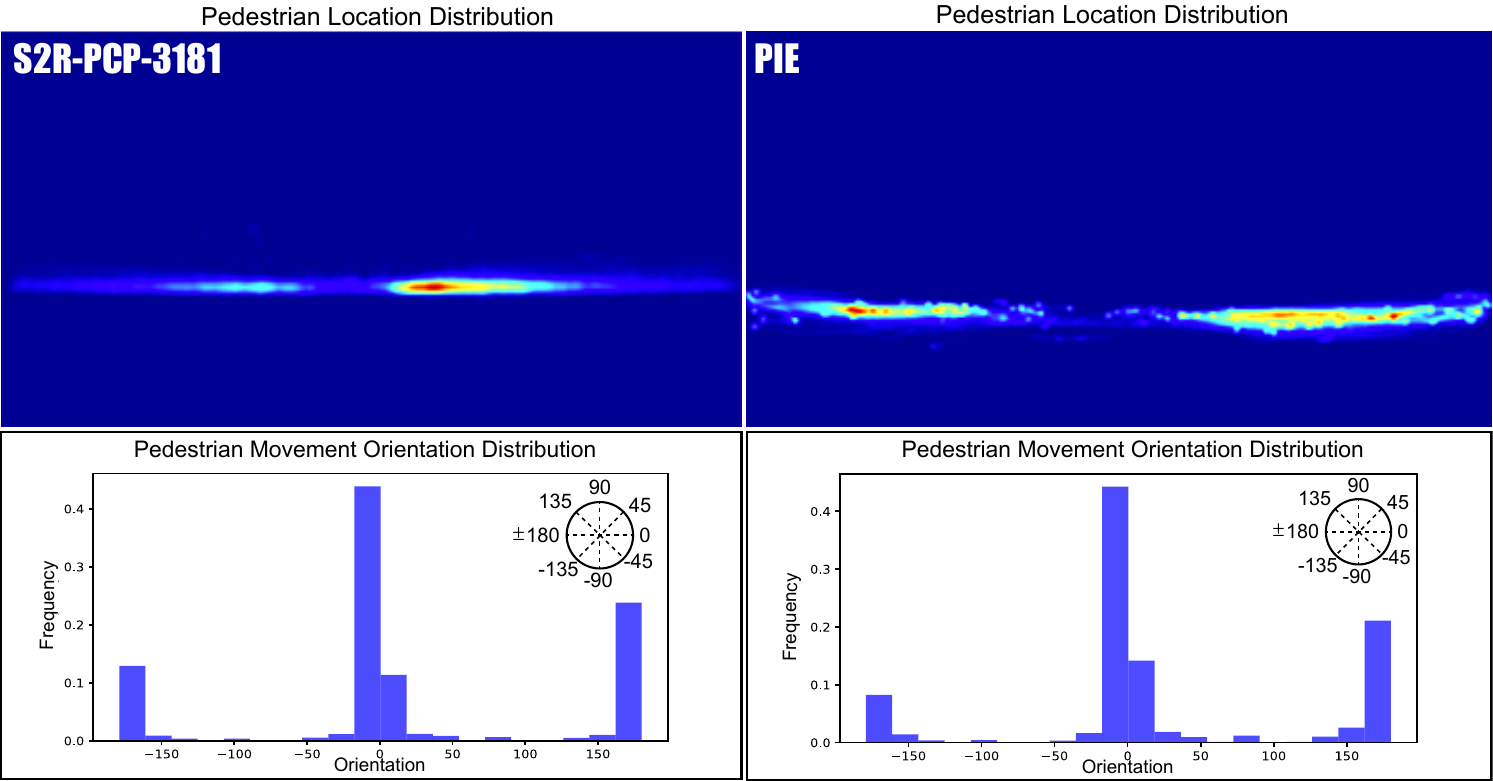}
\caption{The pedestrian location and movement orientation distribution in Syn-PCP-3181 and PIE datasets, where the hotspot region represents the number of pedestrians that pass through these locations.}
\label{fig13}
\end{figure}

\textbf{Real PCP datasets:} JAAD \cite{DBLP:conf/iccvw/RasouliKT17} captures in different countries, which consists of a total of 346 video sequences with a total of 75K frames and 2,786 pedestrians. Among them, 495+191 samples are crossing. The remaining 2,100 pedestrians are not-crossing samples. JAAD has a subset, JAAD\_beh, that includes only 686 behavioral pedestrians. PIE \cite{DBLP:conf/iccv/RasouliKKT19} collects six consecutive hours of daytime data in the city of Toronto, which consists of 55 sequences with 293K frames and 1834 pedestrians. 512 of them are crossing and 1322 are not-crossing samples.

It is known that there is no pedestrian behavior model in CARLA. However, the crossing behavior captured by the ego-view camera often shows relatively horizontal movement. As shown in Fig. \ref{fig13}, the pedestrian location and the movement orientation in Syn-PCP-3181 show similar distributions to PIE. In addition, from the location distribution, the pedestrians in Syn-PCP-3181 own larger distances to the ego-vehicle than the PIE dataset, which indicates that the hidden pedestrian behavior model in Syn-PCP-3181 is complementary and positive for real dataset testing.
Table. \ref{tab1} shows the attribute comparison between S2R-PCP-3181 and other PCP datasets.
  \begin{table}[!t]
	\centering
	\caption{Comparison of S2R-PCP-3181 with other existing pedestrian crossing datasets. Info.: information type of dataset (I: RGB frames; D: depth images; S: semantic images)}
  \setlength{\tabcolsep}{1.2mm}{
		\begin{tabular}{c|c|c|c|c|c}
    \toprule[1.0pt]
    	dataset & \#seqs. & \#frames  & \#peds & Syn./Real. & Info.\\
    	\hline
		JAAD \cite{DBLP:conf/iccvw/RasouliKT17} & 346 & 75K & 686 & Real &  I\\
		PIE \cite{DBLP:conf/iccv/RasouliKKT19} & 55 & 293K & 1800 & Real &  I\\
		Virtual-PedCross-4667 \cite{baijie2022} & 4667 & 745K & 5835 & Syn  & I\\
            S2R-PCP-3181 & 3181 & 490K & 3181 & Syn  & I, D, S\\
    \toprule[1.0pt]
    	\end{tabular}}
	\label{tab1}
\end{table}

\subsection{Implementation Details}

For JAAD and PIE datasets, following the setting of other PCP works, we adopt 16 consecutive video frames (0.5 seconds) as the observation to predict whether pedestrians would cross with the $\tau$ of 30 or 60 frames (1 to 2 seconds). 

In the Knowledge Distiller (Sec. \ref{sec:kd}), the teacher model $\mathcal{T}$ uses 3 default Transformer layers \cite{DBLP:conf/nips/VaswaniSPUJGKP17} with an 8-head self-attention module for each Transformer layer, and generates a 64-dimensional feature vector. The student model $\mathcal{S}$ only has a ResNet18 module and two LSTM layers \cite{DBLP:journals/neco/HochreiterS97}, and each LSTM layer has 100-dimensional hidden states. For pedestrian location information, the input dimension is ($bsize$, 32, 4) of the Knowledge Distiller. For the FOL path in the Knowledge Distiller, the future $P=16$ frames (0.5 seconds) are used.

As stated in the Style Shifter (Sec. \ref{sec:ss}), we crop the local image regions around the pedestrian boxes for the input of the Style Shifter. The local regions are scaled to $112\times112$. Therefore, the shape of all RGB data is ($bsize$, 16, 3, 112, 112), where $bsize=2$ denotes the batch size in all experiments. 
In the Distribution Approximator (Sec. \ref{sec:da}), the input local image region is divided into patches with the size of $16\times16$, which forms the visual tokens in the backbone model $\psi(.)$. The backbone model $\psi(.)$ outputs the feature vector with a dimension of 64. Consequently, 16 frames of observation consist of 784 (112/16$\times$112/16) tokens.

\textbf{Training Details:} The training of teacher model $\mathcal{T}$ and the student model $\mathcal{S}$ takes the Adam optimizer with the learning rate is $10^{-5}$ and a decay of 0.8. The decay step is set to 10, which means that the learning rate is reduced to 80\% in every 10 training epochs. $\mathcal{T}$ is trained with 10 epochs. For the JAAD dataset, the training epoch is set to 40 and the epoch is set to 20 for the PIE dataset. To prevent overfitting, a dropout ratio of 0.5 is introduced. This work is implemented with Python 3.7 and PyTorch 1.7 on a platform with one NVIDIA GeForce RTK 2080Ti GPU and 64 RAM. 
\subsection{Metrics}

Following other PCP works \cite{DBLP:conf/nips/VaswaniSPUJGKP17,DBLP:journals/tits/ZhangAWZ22}, the accuracy (\textbf{Acc}), the area under ROC curve (\textbf{Auc}), precision (\textbf{Pre}), \textbf{Recall} and \textbf{F1}-measure are used to evaluate the models. These metrics all pursue higher values for better performance. 
\subsection{Ablation Studies}
In the ablation studies, we exhaustively evaluate the proposed method's components.

\emph{1) Are the designed syn-to-real knowledge transfer ways suitable for the PCP task?} 

This work designs different syn-to-real knowledge transfer ways for different information. Based on the data form and distribution gaps, we model a \textbf{KnowD} (\emph{Real and Syn}: Ped. Boxes), a \textbf{StyS} (RGB frames), and a \textbf{DistA} (\emph{Syn}: semantic and depth images, \emph{Real}: RGB frames). Whether these kinds of designs take effect on the PCP task? We extensively evaluate different combinations of KnowD, StyS, and DistA. If we do not contain one knowledge transfer module, we can remove the synthetic data path to KnowD, StyS, or DistA, and maintain the real data path, as shown by the blue lines in Fig. \ref{fig2}. If we want to evaluate the performance without all the knowledge transfer modules, we can remove all the synthetic data paths, which means that we only have real data for training.

Table. \ref{tab2} presents the performance on different combinations of KnowD, StyS, and DistA on the PIE dataset \cite{DBLP:conf/iccv/RasouliKKT19}. We can see that KnowD plays an important role in PCP, and each combination with KnowD shows better results than the ones without KnowD. Especially, only KnowD generates the best \textbf{Recall} value in this comparison. The main reason is that the location information in synthetic and real data has the smallest domain gap. Therefore, the promotion role of location clues in synthetic can be directly borrowed. Contrarily, DisA exhibits the worst performance compared with StyS and KnowD, mainly due to the largest distribution gap between semantic or depth images with RGB frames. When fusing all the knowledge transfer ways, the best performance (w/o Recall) is obtained, which means that the Learnable Gated Unit (LGU) can adaptively select the knowledge transfer ways.

\begin{table}[!t]\small
  \centering
  \caption{Performance comparison on the PIE dataset for different knowledge transfer ways, \emph{i.e.}, \textbf{StyS}, \textbf{DistA}, and \textbf{KnowD}.}
    \begin{tabular}{ccc|c|c|c|c|c}
    \toprule[1.0pt]
    StyS& DistA &KnowD&Acc & Auc &F1 & Pre&Rec \\
\hline
   &    &   &  0.81&0.78&0.74&0.70&0.78 \\
 \checkmark   &    &   &  0.82 &  0.74  &  0.62 &   0.67    &0.58 \\
  &  \checkmark  & &  0.79   &  0.71  &  0.57  &  0.63  &  0.53 \\
& &  \checkmark  &  0.88  &  0.86  &  0.80  &  0.76   &   \textbf{0.84}\\
 \checkmark  &\checkmark   && 0.82  & 0.75   & 0.63   &  0.63  & 0.63 \\
\checkmark   &  &\checkmark  & 0.87  &  0.87  &  0.81  &   0.79 &  0.82\\
 & \checkmark  & \checkmark   &  0.87 &  0.86  &  0.79  &      0.78 &  0.80\\
 \checkmark   &  \checkmark  & \checkmark   & \textbf{0.92} &   \textbf{0.90}    &   \textbf{0.82}  &   \textbf{0.83}  & 0.80 \\
    \toprule[1.0pt]
    \end{tabular}%
  \label{tab2}%
\end{table}%

\emph{2) What about the PCP performance by using different domain data in training?} 
\begin{figure}[!t]
\centering
\includegraphics[width=0.9\hsize]{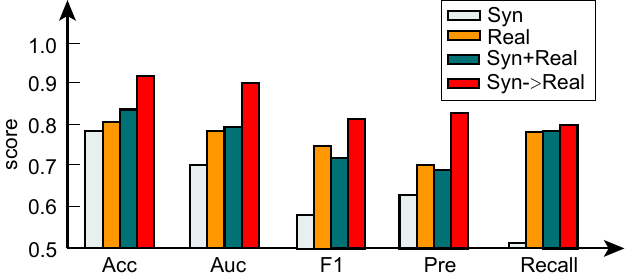}
\caption{Performance comparison on the PIE dataset for ``Syn" training, ``Real" training, ``Syn+Real" training, and our ``Syn$\rightarrow$ Real" training modes.}
\label{fig9}
\end{figure}
After the above evaluation, someone wants to know if we directly combine the synthetic data with the real data together for training, what about the performance? In addition, what about the performance when only using synthetic or real data for training? We seek the answers by comparing the ``\textbf{Syn}" training, ``\textbf{Real}" training, ``\textbf{Syn+Real}" training, and our ``\textbf{Syn$\rightarrow$ Real}" training modes here. As for the ``\textbf{Syn}" training, ``\textbf{Real}" training, and ``\textbf{Syn+Real}" training modes, we only maintain the flowchart marked by blue lines in Fig. \ref{fig2}, \emph{i.e.}, the same flowchart setting as the testing phase without knowledge transfer modules. The testing set of these settings is the same, while the training set is different. ``\textbf{Syn}" training mode inputs the RGB frames and pedestrian locations in synthetic data (3181 sequences). ``\textbf{Real}" is trained on the training data in the PIE dataset, and ``\textbf{Syn+Real}" directly is trained by the combined set of the synthetic data (3181 sequences) and the training set of PIE.

Fig. \ref{fig9} displays the performance difference for these modes. We can see ``\textbf{Syn}" training shows the worst performance because of the domain gap even though it owns the largest training set. ``\textbf{Real}" training provides the comparative Recall value. If we directly combine the synthetic data and real data (``\textbf{Syn+Real}") in the same training way as ``\textbf{Real}",  The F1 and Pre values degrade mainly because of the domain confusion. On the contrary, our formulation, \emph{i.e.}, ``\textbf{Syn$\rightarrow$ Real}", obtains the best performance through the suitable knowledge transfer ways for different information.
\begin{table}[!t]\scriptsize
  \centering
  \caption{Performance influence of the \emph{Backbone} feature model in StyS or DisA, and the \emph{Future Object Location (FOL)} prediction in KnowD on the PIE dataset, where P. (M) means the parameter size and FPS denotes the frames-per-second in the inference stage.}
    \begin{tabular}{l|r|r|r|r|r|r|r}
    \toprule[1.0pt]
\textbf{Backbone}/\textbf{FOL}& \multicolumn{1}{l|}{Acc} & \multicolumn{1}{l|}{Auc} & \multicolumn{1}{l|}{F1} & \multicolumn{1}{l|}{Pre} & \multicolumn{1}{l|}{Rec} & \multicolumn{1}{l|}{P. (M)}& \multicolumn{1}{l}{FPS}\\
\hline
StyS/DistA+\textbf{C3D}\cite{DBLP:conf/iccv/TranBFTP15}   &  0.87     &   0.87    &  0.81     &  0.80   &  \textbf{0.81}&186&52\\
 \verb' '+\textbf{TSM}\cite{DBLP:journals/pami/LinGWH22}  & 0.88   &  0.86    &   0.79  &  0.80  & 0.79&159&57\\
 \verb' '+\textbf{Timesformer}\cite{DBLP:conf/icml/BertasiusWT21} &  \textbf{0.92} &   \textbf{0.90}    &   \textbf{0.82}  &   \textbf{0.83}  & 0.80 &307&30\\
\hline   
    KnowD   & 0.87   &   0.79   &   0.79   &   0.71  & \textbf{0.88} &64&127\\
 \verb' '   \textbf{+FOL}  &  0.88  &  0.86  &  0.80  &  0.76   & 0.84&65&126\\
Gated-S2R-PCP &  0.90     & 0.87     &  0.80  &  0.76   & 0.84&306&30\\
 \verb' '  \textbf{+FOL} &    \textbf{0.92} &   \textbf{0.90}    &   \textbf{0.82}  &   \textbf{0.83}  & 0.80 &307&30\\
    \toprule[1.0pt]
    \end{tabular}%
  \label{tab3}%
\end{table}%

\begin{table*}[!t]\footnotesize
\centering
\caption{Performance comparison of our method and the state-of-the-art on JAAD$_{all}$, JAAD$_{beh}$, and PIE datasets. The best and second values for each metric are marked by \textcolor{red}{red} and \textcolor{blue}{blue} color, respectively.}
\begin{tabular}{ccccccc|rrrrr|ccccc}
  \toprule[1.0pt]
\multicolumn{ 2}{c|}{Model} &   \multicolumn{5}{c|}{JAAD$_{all}$} &  \multicolumn{5}{c|}{JAAD$_{beh}$} & \multicolumn{5}{c}{PIE} \\
\cline{3-17}
\multicolumn{ 2}{c|}{} & Acc & Auc &  F1 &  Pre &  Rec & Acc & Auc & F1 & Pre & Rec & Acc & Auc & F1 & Pre & Rec \\
\hline
\multicolumn{2}{c|}{ATGC \cite{DBLP:conf/iccvw/RasouliKT17} $_{\text{ICCVW}2017}$} & 0.67 & 0.62 & 0.76 & 0.72 & 0.80 & 0.48 & 0.41 & 0.62 & 0.58 & 0.66 & 0.59 & 0.55 & 0.39 & 0.33 & 0.47 \\

\multicolumn{2}{c|}{StackedRNN \cite{DBLP:conf/cvpr/NgHVVMT15}$_{\text{CVPR}2015}$} & 0.79 & 0.79 & 0.58 & 0.46 &	0.79 & 0.60 & 0.60 & 0.66 & \textcolor{blue}{0.73} & 0.61 & 0.82 & 0.78	& 0.67 & 0.67 & 0.68\\

\multicolumn{2}{c|}{MultiRNN \cite{DBLP:conf/cvpr/BhattacharyyaFS18}$_{\text{CVPR}2018}$} & 0.79 & 0.79 & 0.58 & 0.45 & 0.79 & 0.61 & 0.50 & 0.74 & 0.64 & 0.86 & 0.83 & 0.80 & 0.71 & 0.69 & 0.73 \\

\multicolumn{2}{c|}{SFGRU \cite{DBLP:conf/bmvc/RasouliKT19}$_{\text{BMVC}2019}$} & 0.83 & 0.77 & 0.58 & 0.51 & 0.67 & 0.58 & 0.56 & 0.65 & 0.68 & 0.62 & 0.86 & 0.83 & 0.75 & 0.73 & 0.77 \\

\multicolumn{2}{c|}{SingleRNN-GRU \cite{DBLP:conf/ivs/KotserubaRT20}$_{\text{IV}2020}$} & 0.65 & 0.59 & 0.34 & 0.26 & 0.49 & 0.58 & 0.54 & 0.67 & 0.67 & 0.68 & 0.83 & 0.77 & 0.67 & 0.70 & 0.64 \\

\multicolumn{2}{c|}{SingleRNN-LSTM \cite{DBLP:conf/ivs/KotserubaRT20}$_{\text{IV}2020}$}  & 0.78 & 0.75 & 0.54 & 0.44 & 0.70 & 0.51 & 0.48 & 0.61 & 0.63 & 0.59 & 0.81 & 0.75 & 0.64 & 0.67 & 0.61 \\

\multicolumn{2}{c|}{PCPA \cite{DBLP:conf/wacv/KotserubaRT21}$_{\text{WACV}2021}$} & 0.83 & 0.83 & 0.64 & - & - & 0.56 & 0.56 & 0.68 & - & - & 0.87 & 0.86 & 0.77 & - & - \\

\multicolumn{2}{c|}{BiPed \cite{DBLP:conf/iccv/RasouliRL21}$_{\text{ICCV}2021}$}  & 0.83 & 0.79 & 0.60 & 0.52 & - & -	& - & - & -	& - & \textcolor{blue}{0.91} & \textcolor{blue}{0.90} & \textcolor{red}{0.85} & \textcolor{blue}{0.82} & -	\\

\multicolumn{2}{c|}{IntFormer \cite{DBLP:journals/corr/abs-2105-08647}$_{2021}$} & 0.86 & 0.78 & 0.62 & - & - & 0.59 & 0.54 & 0.69 & - & - & 0.89 & \textcolor{red}{0.92} & 0.81 & - & -\\

\multicolumn{2}{c|}{ST-CrossingPose \cite{DBLP:journals/tits/ZhangAD22}$_{\text{TITS}2022}$} & - & - & - & - & - & 0.63 & 0.56 & 0.74 & 0.66 & 0.83 & - & - & -  & - & - \\

\multicolumn{2}{c|}{Yang \emph{et al.} \cite{DBLP:journals/tiv/YangZYRO22}$_{\text{TIV}2022}$} & 0.83 & 0.82 & 0.63 & 0.51 & \textcolor{blue}{0.81} & 0.62 & 0.54 & 0.74 & 0.65	& \textcolor{blue}{0.85} & 0.89 & 0.86 & 0.80 & 0.79 & \textcolor{blue}{0.80}\\

\multicolumn{2}{c|}{Pedestrian Graph +(32) \cite{DBLP:journals/tits/CadenaQWY22}$_{\text{TITS}2022}$} &0.86	& \textcolor{red}{0.88} & 0.65 & 0.58 & 0.75	& \textcolor{red}{0.70} & \textcolor{red}{0.70} &0.76 & \textcolor{red}{0.77} & 0.75 & 0.89 & \textcolor{blue}{0.90} & 0.81 & \textcolor{red}{0.83} & 0.79 \\
\multicolumn{2}{c|}{VR-PCP \cite{baijie2022}$_{\text{ITSC}2022}$} & 0.86 & 0.81 & \textcolor{blue}{0.77} & \textcolor{blue}{0.74} & 0.81 & 0.64 & 0.66 & 0.76 & 0.70 & \textcolor{red}{0.89} & 0.89 & 0.88 & 0.67 & 0.74 & \textcolor{red}{0.84} \\
\multicolumn{2}{c|}{Dong \cite{dong2023pedestrian}$_{\text{ICLRW}2023}$} & \textcolor{blue}{0.88} & 0.81 & 0.69&-& - & 0.64 & 0.56 &0.75 &-& - & 0.89 & 0.88 & 0.79 & - & - \\
\multicolumn{2}{c|}{PIT* \cite{DBLP:journals/tits/ZhouTZLG23}$_{\text{IEEE-TITS}2023}$} & 0.86&  \textcolor{blue}{0.87} & 0.66 &0.55& \textcolor{red}{0.82} &  \textcolor{blue}{0.69} & 0.67 & \textcolor{red}{0.78} & \textcolor{blue}{0.74} &0.83 & 0.90 &  \textcolor{red}{0.90} & 0.80 & 0.81 &\textcolor{blue}{0.80} \\
\multicolumn{2}{c|}{Pedformer \cite{DBLP:conf/icra/RasouliK23}$_{\text{ICRA}2023}$} & \textcolor{red}{0.93} & 0.76 &0.54 & 0.65 & - & -& -& -& - &-& -& -& -& - &- \\
\hline
\multicolumn{2}{c|}{Gated-S2R-PCP (Ours)} & \textcolor{blue}{0.88} & 0.84   & \textcolor{red}{0.79} & \textcolor{red}{0.77} & \textcolor{red}{0.82} & 0.66 & \textcolor{blue}{0.68} & \textcolor{blue}{0.77} & 0.71 & \textcolor{blue}{0.85} & \textcolor{red}{0.92} & \textcolor{blue}{0.90} & \textcolor{blue}{0.82} & \textcolor{red}{0.83} &  \textcolor{blue}{0.80} \\

\toprule[1.0pt]
\end{tabular}
   \begin{tablenotes}
\item \scriptsize{*PIT has four method versions and the best one for certain metrics in JAAD and PIE are different. Therefore, we take the average metric value of method versions here. }
\end{tablenotes}
\label{tab4}
\end{table*}

\emph{3) Which backbone $\psi(.)$ is better for StyS or DistA and how about the influence when adding the Future Object Location (FOL) prediction task in KnowD?} 

\textbf{Backbone Sensitivity:} In the StyS and DistA, we introduce a backbone model for feature extraction, which may influent the final performance. We evaluate the role of different backbones here. We replace the Timesformer \cite{DBLP:conf/icml/BertasiusWT21} with 3D Convolution (C3D) \cite{DBLP:conf/iccv/TranBFTP15} and Temporal Shift Model (TSM) \cite{DBLP:journals/pami/LinGWH22}, respectively, both of which are commonly used video coding networks. C3D is pre-trained on the UCF101 dataset \cite{DBLP:journals/corr/abs-1212-0402}. TSM is extended from the pre-trained ResNet 50 \cite{DBLP:conf/cvpr/HeZRS16}. Timesformer \cite{DBLP:conf/icml/BertasiusWT21} here is not pre-trained and directly used for training. The top half of Table. \ref{tab3} displays the results with different backbones in StyS and DistA. From the results, we can see that Timesformer shows the best performance. However, C3D and TSM are also comparative. Therefore, the self-attention mechanism in Timesformer plays a positive role. Additionally, we also list the parameter size and inference efficiency here. It shows that the backbone model Timesformer has a more expensive computation cost than TSM and C3D. However, Timesformer boosts the PCP performance and meets the real-time demand.

\textbf{FOL Influence:} We also evaluate the influence of FOL in the KnowD module which forms multi-task learning for location prediction and crossing prediction. Here, we provide two groups: only KnowD for training (KnowD with or without FOL) and the whole model for training (Gated-S2R-PCP with or without FOL). The results are listed in the bottom half of the Table. \ref{tab3}, and show that adding FOL can improve the performance except for the Recall value, and the pure \textbf{KnowD-w/o-FOL} obtains 0.88 of Recall value. The reason is that location information has the closest relation to the crossing behavior with large displacement. When fusing FOL and PCP, FOL relies on the historical motion pattern in the observation (may with small displacement), which may enforce a motion-consistent constraint on the PCP and show weak adaptation to the large-motion in crossing. In Sec. \ref{sec:longFOL}, we will evaluate the influence of FOL with different prediction windows.
\begin{figure}[!t]
\centering
\includegraphics[width=\hsize]{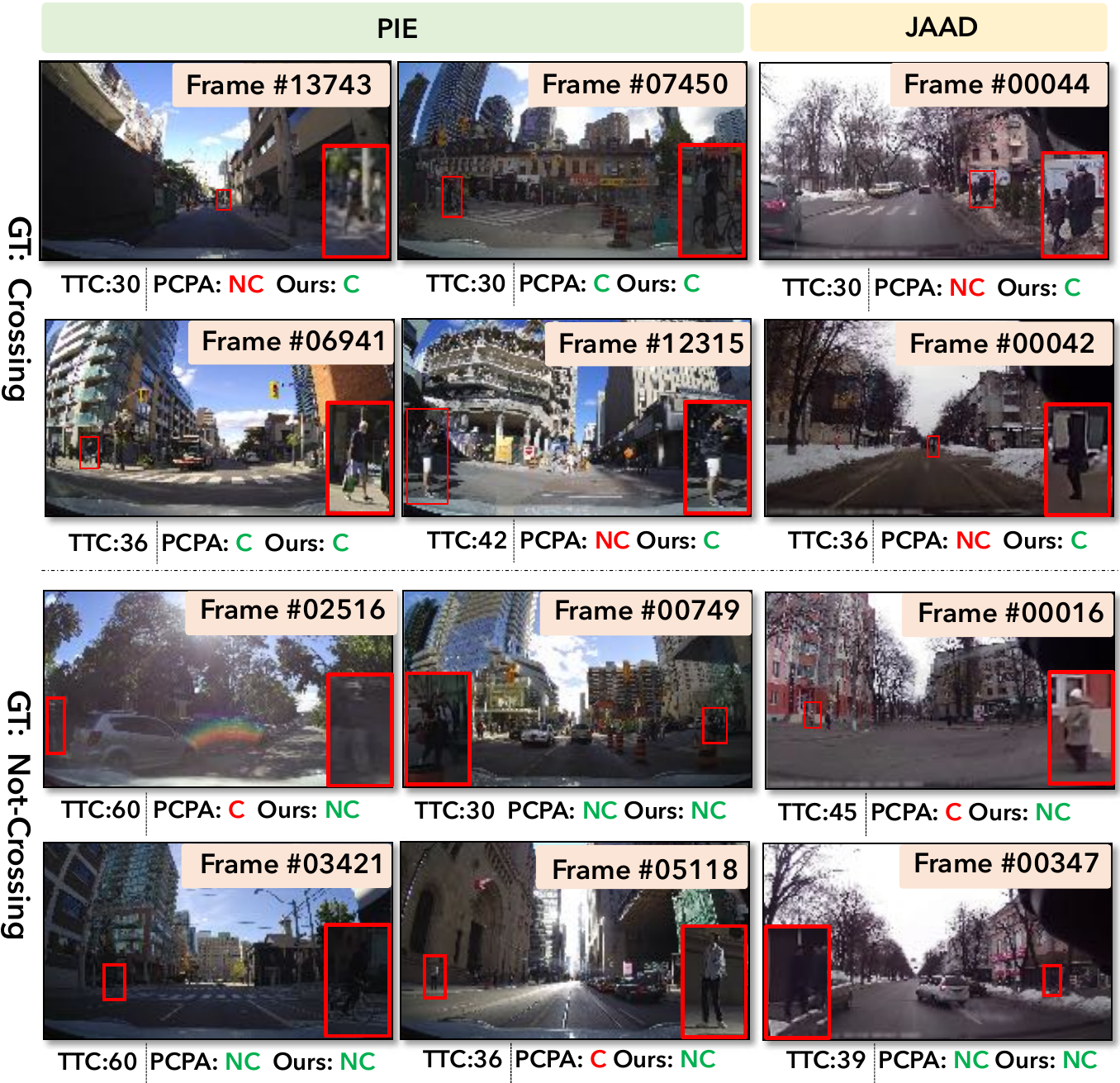}
\caption{Visualization comparison of some examples between PCPA \cite{DBLP:conf/wacv/KotserubaRT21} and our method, where GT is ground truth, C is Crossing, and NC is not crossing. TTC denotes the video frames of time-to-crossing.}
\label{fig10}
\end{figure}
\subsection{Comparison with SOTA Methods}

To verify the proposed method, we take three datasets, i.e., JAAD$_{all}$, JAAD$_{beh}$, and PIE, in the evaluation, and compare it with sixteen state-of-the-art methods. They are ATGC \cite{DBLP:conf/iccvw/RasouliKT17} , StackedRNN \cite{DBLP:conf/cvpr/NgHVVMT15}, MultiRNN \cite{DBLP:conf/cvpr/BhattacharyyaFS18}, SFGRU \cite{DBLP:conf/bmvc/RasouliKT19}, SingleRNN-GRU \cite{DBLP:conf/ivs/KotserubaRT20}, SingleRNN-LSTM \cite{DBLP:conf/ivs/KotserubaRT20}, PCPA \cite{DBLP:conf/wacv/KotserubaRT21},  BiPed \cite{DBLP:conf/iccv/RasouliRL21}, IntFormer \cite{DBLP:journals/corr/abs-2105-08647}, ST-CrossingPose \cite{DBLP:journals/tits/ZhangAD22}, Pedestrian Graph+(32)\cite{DBLP:journals/tits/CadenaQWY22}, VR-PCP \cite{baijie2022}, PIT \cite{DBLP:journals/tits/ZhouTZLG23}, Pedformer \cite{DBLP:conf/icra/RasouliK23}, the methods proposed by Yang \emph{et al.} \cite{DBLP:journals/tiv/YangZYRO22} and Dong \cite{dong2023pedestrian}, respectively. The results are listed in Table. \ref{tab4}.

From the results, we can observe that our Gated-S2R-PCP obtains the comparative performance for all the metrics in three datasets. It seems that Pedestrian Graph+(32) \cite{DBLP:journals/tits/CadenaQWY22}, our Gated-S2R-PCP, PIT \cite{DBLP:journals/tits/ZhouTZLG23}, and VR-PCP \cite{baijie2022} are the top four methods for JAAD dataset. 
Pedestrian Graph+(32) adopts 32 frames as the observation and the graph relation between pedestrians is considered. VR-PCP also leverages the knowledge of synthetic data for the PCP task, where only the RGB frames and pedestrian boxes are used. PIT fulfills the Progressive Transformer for PCP, where it designs 15 transformer layers with the inputs of RGB frames, pedestrian poses, and ego-vehicle motion state. Notably, Pedformer generates 0.93 accuracy in the JAAD$_{all}$ dataset, while the other metrics are not promising.

For the PIE dataset, the top four ones are Gated-S2R-PCP, PIT, BiPed \cite{DBLP:conf/iccv/RasouliRL21}, and Pedestrian Graph+(32) \cite{DBLP:journals/tits/CadenaQWY22}. BiPed considers the trajectory prediction in the PCP task, and multi-task learning is formulated. In addition, a categorical interaction between pedestrians is also modeled in BiPed, which is promising for the PCP task in complex urban scenes. These observations indicate that more input frames and attributes will boost the performance, and future trajectory guidance can reduce the determination uncertainty of pedestrian crossing. 

In addition, we also compare the performance of our Gated-S2R-PCP and PCPA \cite{DBLP:conf/wacv/KotserubaRT21} with several snapshots for the frames at time $T+\tau$ (\emph{i.e.}, crossing or not-crossing intention frames) in Fig. \ref{fig10}. From this figure, we can see that the small-scale issue of pedestrians has a more negative impact on PCPA than our method, and our Gated-S2R-PCP shows more consistent results to the ground truth.

\subsection{Further PCP Analysis on Near-Collision Scenes}
To verify Gated-S2R-PCP for the extra pedestrian crossing scenarios, this work takes all pedestrian crossing sequences (50 ones) in the challenging  DADA-2000 \cite{DBLP:journals/tits/FangYQXY22} dataset (\emph{abbrev.,} \textbf{PC50-DADA}), where each crossing pedestrian is hit by the ego-vehicle. Certainly, for further PCP analysis, we can also implement the proposed method to the practical perception system of self-driving vehicles by Tensor RT or some other platforms. However, it needs to pre-detect and track the pedestrians, which involves perception bias for the PCP evaluation. Therefore, we instead take the near-collision scenes here for an in-depth analysis of our Gated-S2R-PCP, to be with a closer relation to the safe-driving function than ever before.

\begin{figure}[!t]
\centering
\includegraphics[width=0.98\hsize]{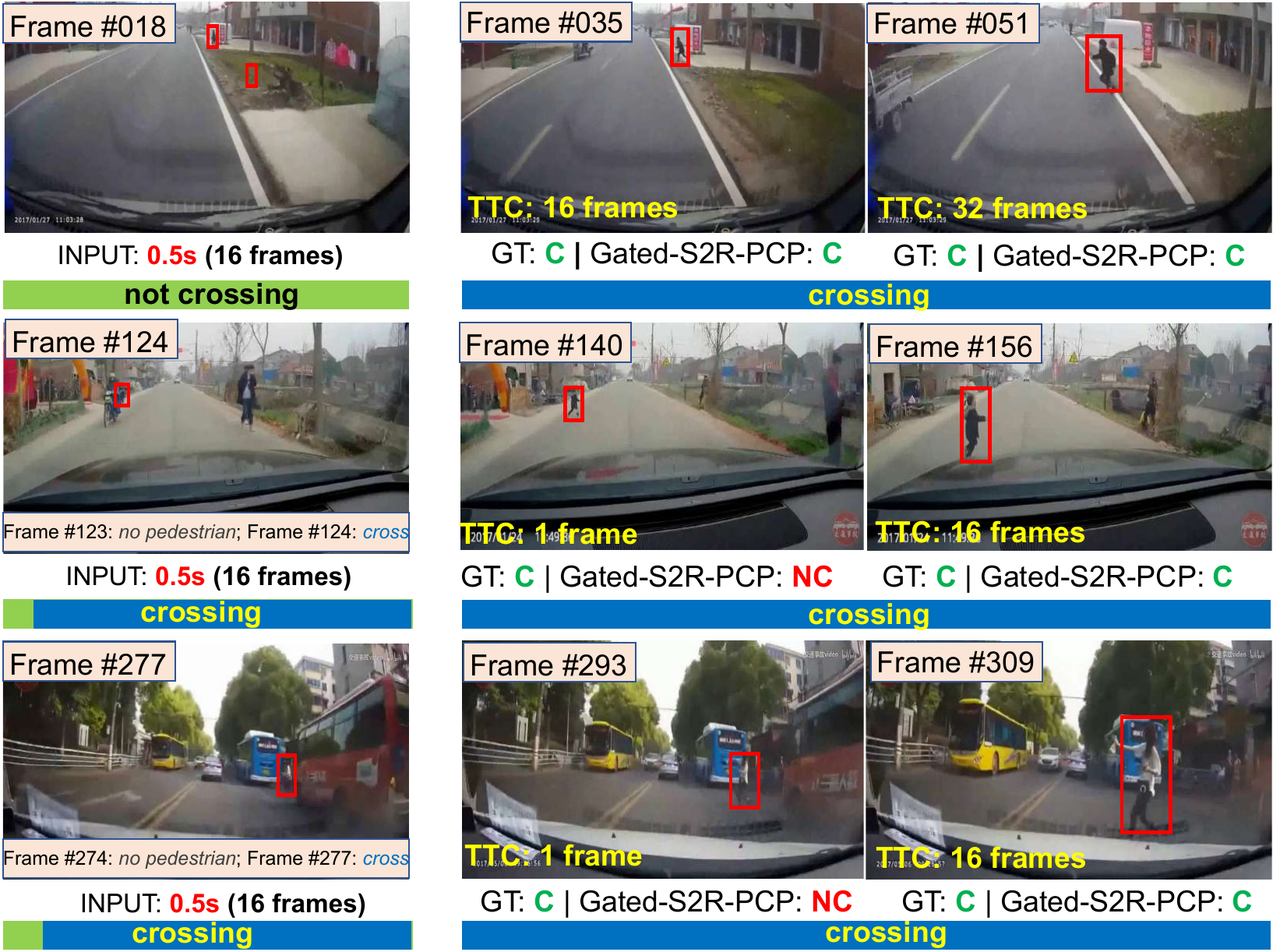}
\caption{The visualization of PCP results in PC50-DADA, where the red boxes mark the pedestrians whose crossing intention is to be predicted.}
\label{fig14}
\end{figure}

Different from the JAAD or PIE datasets, the pedestrians in $86\%$ sequences of PC50-DADA appear suddenly and begin to cross the road without enough Time-to-Crossing (TTC) interval (\emph{e.g.}, 1-2 seconds). Hence, for the $86\%$ sudden crossing sequences, we postpone the time $T+\tau$ to adapt to the input observation setting of $T=16$ frames of PCP. As shown in the second and third rows of Fig. \ref{fig14}, the observation $T$ (16) frames contain the crossing pedestrians. Additionally, we manually annotate 100 not-crossing pedestrians in PC50-DADA to serve the negative samples. With the sudden crossing situation, we set TTC $\tau$ from 1 to 16 frames (\emph{i.e.}, 0.03-1 seconds) in this evaluation.

Table. \ref{tab5} lists the results with varying TTC $\tau$. The performance of Gated-S2R-PCP increases when the pedestrian is close to the ego-vehicle, which meets our expectation because a larger TTC means that the pedestrian at time $T+\tau$ has a higher collision risk. Fig. \ref{fig14} also showcases the PCP results in three samples of PC50-DADA. Notably, the second and third rows of PCP results generate failure predictions at time $T+1$ and re-obtain accurate results for $T+16$. This observation indicates that PCP models prefer the large pedestrian appearance and location distances at time $T$ and time $T+\tau$, and the near-collision involved PCP task has a large space to be improved.
\begin{table}[!t]\small
  \centering
  \caption{PCP Performance of the proposed method in PC50-DADA.}
    \begin{tabular}{c|c|c|c|c|c|c||c}
    \toprule[1.0pt]
TTC $\tau$&1.0s&0.8s&0.6s&0.4s&0.2s&0.03s&all $\tau$\\
\hline
Acc &  0.63 & 0.60  &  0.57 & 0.58& 0.57&0.53&0.66\\
Auc& 0.65&0.62 &0.61 & 0.59 &0.63&0.58&0.68\\
    \toprule[1.0pt]
    \end{tabular}%
  \label{tab5}%
\end{table}%

\subsection{Discussions}
\emph{1) Is the shared feature space obtained after DisA?}
\begin{figure}[!t]
\centering
\includegraphics[width=\hsize]{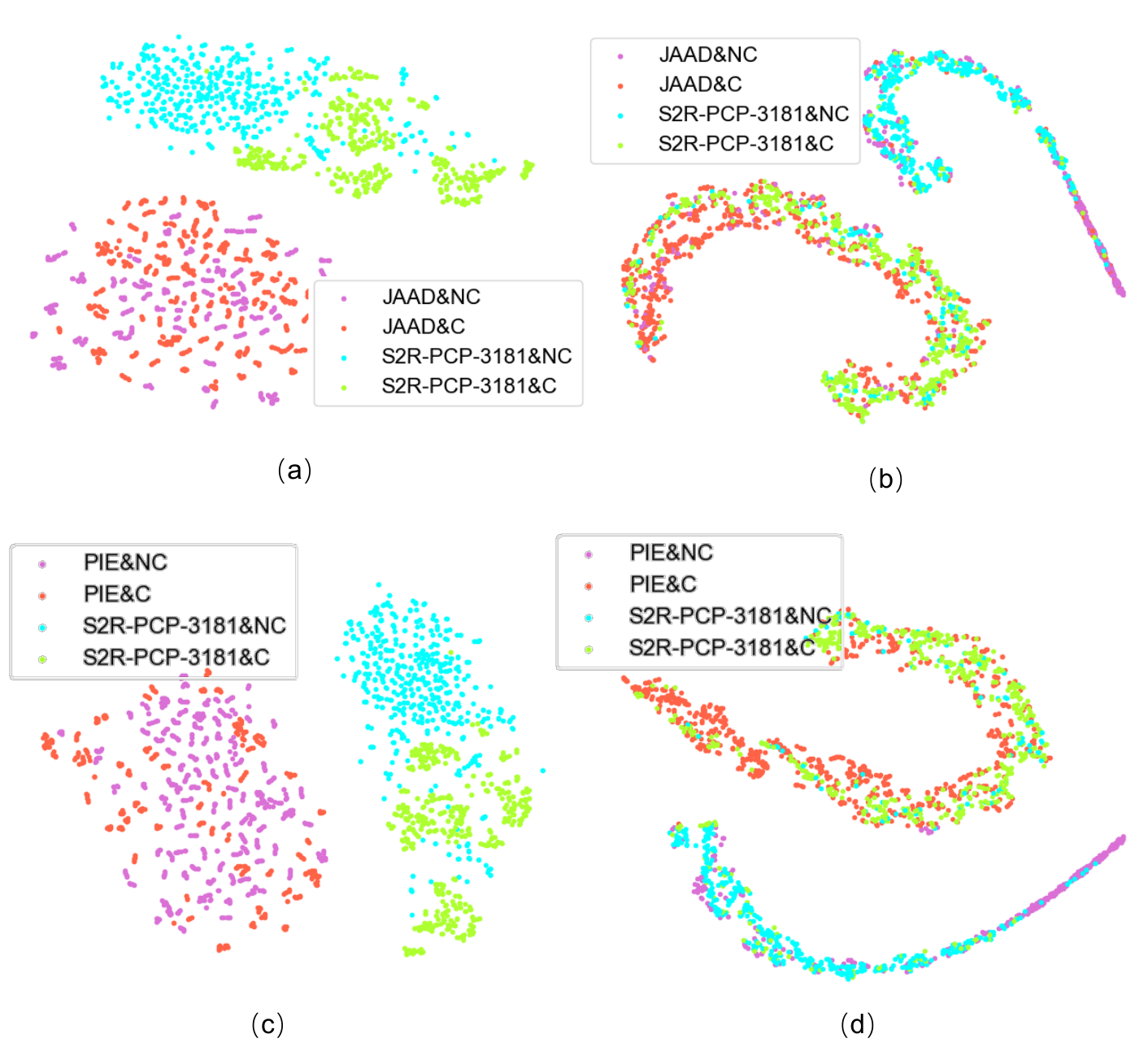}
\caption{Feature distributions of \emph{Before-DistA} and \emph{After-DistA} for pedestrian crossing and not-crossing using t-SNE \cite{van2008visualizing} with 1000 samples from S2R-PCP-3181 or real datasets (taking 500 crossing and 500 not-crossing samples in each dataset). (a) and (c) denote the feature distributions of \emph{Before-DistA}. (b) and (d) specify the feature distributions of \emph{After-DistA}.}
\label{fig11}
\end{figure}

In this work, we model a Distribution Approximator (DistA) for the large domain gap between semantic and depth images in synthetic data and the RGB frames in real data. Is the shared feature space obtained after DistA? We verify the effect of DistA by visualizing the feature point distribution using t-SNE \cite{van2008visualizing}.
Specifically, we randomly select 1000 samples (crossing or not-crossing clips) in the synthetic S2R-PCP-3181 dataset and the real JAAD or PIE dataset. Here, we adopt the Timesformer \cite{DBLP:conf/icml/BertasiusWT21} as the feature backbone model and generate a 512-dimensional vector for the observation (16 frames of 0.5 seconds) of each sample. We use t-SNE to reduce the feature dimension, \emph{i.e.}, $512\rightarrow 2$. Then the feature vectors are visualized in Fig. \ref{fig11}, where (a) and (c) are the feature distribution before DistA.  (b) and (d) are the feature distribution after DistA.

The visualization shows that the confused crossing and not-crossing feature points are separated from the structured manifolds. Notably, from the feature distribution before DistA, we can see that the synthetic samples (semantic and depth images) have more clearer margin for crossing and not-crossing behavior than the real data (RGB frames). Therefore, synthetic data's behavior knowledge can help guide the distinction of behavior patterns in real data via the distribution approximating process.
In addition, there is a long-tailed distribution for the not-crossing samples, which may be caused by the more confused motion pattern compared with the crossing behaviors. Through this evaluation, we find the shared feature space between crossing or not-crossing in synthetic and real data obtained by the DistA.

\begin{figure}[!t]
\centering
\includegraphics[width=\hsize]{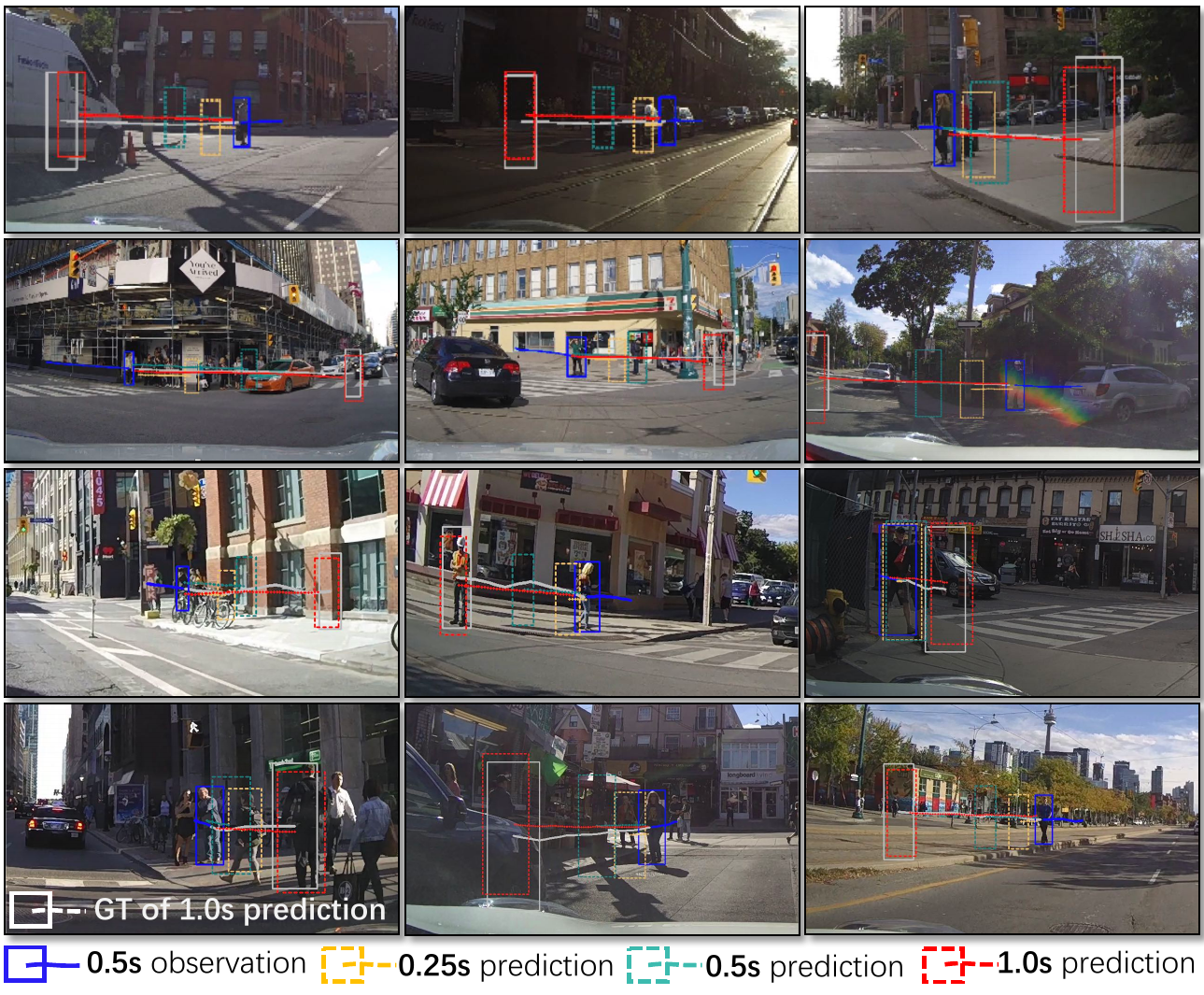}
\caption{Some prediction results of Future Object Location (FOL) with time window (0.25s, 0.5s, and 1.0s) on the PIE dataset.}
\label{fig12}
\end{figure}

\emph{2) What about the interactive role between FOL and PCP?}
\label{sec:longFOL}
As evaluated in the ablation studies, the FOL prediction has a positive role for PCP except for the Recall value. To further check the reason, we evaluate the interactive influence between FOL and PCP here. As for FOL, we adopt the Average Intersection of Unit (\textbf{AIoU},\%) and Final Intersection of Unit (\textbf{FIoU},\%) between the predicted bounding boxes and the ground truth. FIoU is computed by the predicted location in the last frame while AIoU focuses on all the prediction windows. Similarly, the trajectory point errors (Euclidean distance between the predicted trajectory points with the ground-truth), \emph{i.e.}, Average Distance Error (\textbf{ADE}, \emph{pixels}) and Final Distance Error (\textbf{FDE}, \emph{pixels}), are also utilized. The results are displayed in Table. \ref{tab6}. We also present some prediction results of FOL with different time windows in Fig. \ref{fig12}.

We can observe that 0.5s of FOL prediction can obtain the best PCP performance for Acc, Auc, and Pre values. The Recall value may be weakened by long-term FOL mainly because of the inconsistent behavior type in observation (not-crossing or wandering) and crossing in prediction. In addition, we can see FOL can provide a clear separation between crossing and not-crossing behaviors, as shown in the top-right image in Fig. \ref{fig12}. The woman may cross but the future location presents a clear not-crossing behavior along the roadside. Longer-term FOL generates weaker trajectory prediction performance (with larger ADE and FDE). In common sense, AIoU and FIoU are larger for the shorter-term FOL prediction, and vice versa. However, there is one interesting phenomenon the AIoU and FIoU of 0.5s are larger than the ones of 0.25s. The reason for this is that when predicting the FOL in a short time, the small pedestrian displacement has little influence on the FOL loss computation in training. When increasing the prediction window, the consistency of the movement direction has an important impact on the FOL loss computation in prediction. Besides, longer-term FOL prediction may involve a large loss value. Therefore, AIoU and FIoU increase from 0.25s to 0.5s predictions and decrease from 0.5s to 1.0s predictions.
\begin{table}[!t]\footnotesize
  \centering
  \caption{Performance influence PCP and FOL (with 0.25s (8 frames), 0.5s (16 frames), and 1.0s (32 frames) prediction) on the PIE dataset, where $\uparrow$ and $\downarrow$ prefer small and large value, respectively.}
  \setlength{\tabcolsep}{1.2mm}{
    \begin{tabular}{r|r|r|r|r|r||r|r|r|r}
    \toprule[1.0pt]
\textbf{Pred. Set}& \multicolumn{1}{l|}{Acc} & \multicolumn{1}{l|}{Auc} & \multicolumn{1}{l|}{F1} & 
\multicolumn{1}{l|}{Pre} & 
\multicolumn{1}{l||}{Rec} & \multicolumn{1}{l|}{AIoU$\uparrow$} & \multicolumn{1}{l|}{FIoU$\uparrow$} & \multicolumn{1}{l|}{ADE$\downarrow$} & 
\multicolumn{1}{l}{FDE$\downarrow$} \\
\hline
0.25s (8-f.)  & 0.89   &   0.89   &   \textbf{0.83}   & 0.81 & \textbf{0.84} & 44.17  & 24.06 & \textbf{20.78} &\textbf{23.44} \\
0.5s (16-f.)  &  \textbf{0.92}  &  \textbf{0.90} &  0.82  & \textbf{0.83} & 0.80 & \textbf{47.72}   & \textbf{42.13} & 22.09&27.72\\
1.0s (30-f.) & 0.88   & 0.87  &  0.80  & 0.81 & 0.78 & 39.14   & 24.04 & 44.05& 68.50\\
    \toprule[1.0pt]
    \end{tabular}}%
  \label{tab6}%
\end{table}%

\section{Conclusions}
\label{con}

Inspired by the conventional wisdom of divider-and-conquer, this paper proposes the Gated Syn-to-Real knowledge transfer model for Pedestrian Crossing Prediction (Gated-S2R-PCP). Three knowledge transfer ways, \emph{i.e.}, a Style Shifter, a Distribution Approximator, and a Knowledge Distiller, are designed to transfer the visual context, semantic and depth information, and the locations of pedestrians in synthetic data (our S2R-PCP-3181) to two real datasets (JAAD and PIE). A Learnable Gated Unit (LGU) is modeled to adaptively fuse knowledge features for PCP. Through extensive experiments and further PCP evaluation on near-collision scenes in the DADA-2000 dataset, the superiority of Gated-S2R-PCP is verified. Besides, we also evaluate the interactive role of FOL prediction with PCP, and 0.5s of FOL prediction is the best choice for our PCP task. 

From the investigation of this work, some highlights can be concluded. (1) A differentiated knowledge transfer from synthetic to real data is promising when meeting different information domain gaps. (2) The extra Future Object Location (FOL) prediction task benefits the precision and accuracy of PCP. (3) The PCP task in the near-collision scenes is challenging and with large space to be improved.

{\small
\bibliographystyle{IEEEtran}
\bibliography{ref}
}

\end{document}